\documentclass[journal]{IEEEtran}
\ifCLASSINFOpdf
  \usepackage[pdftex]{graphicx}
\else
\fi

\usepackage{mathrsfs}
\usepackage{multirow}
\usepackage{arydshln}
\usepackage{booktabs}
\usepackage{hyperref}
\usepackage{color}
\usepackage{amsmath, bm}
\usepackage[ruled,linesnumbered]{algorithm2e}
\usepackage{tabu}
\usepackage{cancel}
\usepackage{subfig}
\usepackage[table]{xcolor}
\usepackage{fp}  

\usepackage{flushend}

\usepackage{mathrsfs}
\usepackage{adjustbox}

\ifCLASSINFOpdf
\else
\fi
\usepackage{amssymb}
\usepackage{mathrsfs}
\usepackage{amsmath}
\usepackage{bm}
\usepackage{makecell}
\usepackage{multirow}
\usepackage{colortbl}
\usepackage{soul}
\usepackage{caption}
\usepackage[figuresleft]{rotating}
\definecolor{chromeyellow}{rgb}{1.0, 0.65, 0.0}
\definecolor{carmine}{rgb}{0.59, 0.0, 0.09}
\usepackage[percent]{overpic}

\begin{document}

\captionsetup[table]{skip=0pt}
%
\title{Towards Unified Semantic and Controllable Image Fusion: A Diffusion Transformer Approach}
%
%
%

\author{Jiayang Li$^{\ast}$,
        Chengjie Jiang$^{\ast}$,
        Junjun~Jiang$^{\dagger}$,~\IEEEmembership{Senior Member,~IEEE},
        Pengwei Liang,
        Jiayi Ma,~\IEEEmembership{Senior Member,~IEEE},
        and Liqiang~Nie,~\IEEEmembership{Senior Member,~IEEE}


\thanks{J. Li, P. Liang and J. Jiang are with the Faculty of Computing, Harbin Institute of Technology, Harbin 150001. E-mail: lijiayang.cs@gmail.com, erfect2020@gmail.com, jiangjunjun@hit.edu.cn.}
\thanks{C. Jiang is with Tsinghua Shenzhen International Graduate School, Tsinghua University, Shenzhen, China. E-mail: 18601580580@163.com.}
\thanks{J. Ma is with the Electronic Information School, Wuhan University, Wuhan 430072, China. E-mail: jyma2010@gmail.com.}
\thanks{L. Nie is with the School of Computer Science and Technology, HarbinInstitute of Technology (Shenzhen), Shenzhen 518055, China. E-mail:{nieliqiang@gmail.com}.}
\thanks{$^{\ast}$Jiayang Li and Chengjie Jiang contributed equally to this work.}
\thanks{$^{\dagger}$Corresponding author: Junjun Jiang.}
}

\markboth{Journal of \LaTeX\ Class Files,~Vol.~14, No.~8, August~2021}%
{Shell \MakeLowercase{\textit{et al.}}: Bare Demo of IEEEtran.cls for IEEE Journals}
%



\maketitle

\begin{abstract}

Image fusion aims to blend complementary information from multiple sensing modalities, yet existing approaches remain limited in robustness, adaptability, and controllability. Most current fusion networks are tailored to specific tasks and lack the ability to flexibly incorporate user intent, especially in complex scenarios involving low-light degradation, color shifts, or exposure imbalance. Moreover, the absence of ground-truth fused images and the small scale of existing datasets make it difficult to train an end-to-end model that simultaneously understands high-level semantics and performs fine-grained multimodal alignment. We therefore present DiTFuse, instruction-driven Diffusion–Transformer (DiT) framework that performs end-to-end, semantics-aware fusion within a single model. By jointly encoding two images and natural-language instructions in a shared latent space, DiTFuse enables hierarchical and fine-grained control over fusion dynamics, overcoming the limitations of pre-fusion and post-fusion pipelines that struggle to inject high-level semantics. The training phase employs a multi‑degradation masked‑image modeling strategy, so the network jointly learns cross‑modal alignment, modality‑invariant restoration, and task‑aware feature selection without relying on ground truth images. A curated, multi‑granularity instruction dataset further equips the model with interactive fusion capabilities. DiTFuse unifies infrared–visible, multi‑focus, and multi‑exposure fusion—as well as text‑controlled refinement and downstream tasks—within a single architecture. Experiments on public IVIF, MFF, and MEF benchmarks confirm superior quantitative and qualitative performance, sharper textures, and better semantic retention. The model also supports multi‑level user control and zero‑shot generalization to other multi‑image fusion scenarios, including instruction‑conditioned segmentation. The code is available
at \href{https://github.com/Henry-Lee-real/DiTFuse}{https://github.com/Henry-Lee-real/DiTFuse}.
\end{abstract}


%
\IEEEpeerreviewmaketitle

\section{Introduction}\label{sec1}


Due to the inherent performance bottlenecks of hardware and the complexity of the perceived environment, a single imaging modality can only capture partial information of a natural scene. Image fusion technology integrates complementary information from multiple sources to generate a fused image with more comprehensive information. Depending on the specific task, image fusion techniques mainly include infrared–visible image fusion (IVIF), multi-focus fusion (MFF), and multi-exposure fusion (MEF). These fusion technologies are widely used in mobile photography~\cite{intro_mff1,intro_mff2,UltraFusion}, autonomous driving~\cite{intro_mef2}, and medical imaging~\cite{intro_mef3}, where they play a crucial role in enhancing scene perception and improving visual effects.

Although existing methods have demonstrated excellent visual effects in their fusion results, they often fail to significantly improve the accuracy of downstream tasks that use these fused images. To enhance both the visual perceptual quality of the fusion results and the performance of downstream tasks simultaneously, much current work injects high-level semantic information into the fusion network in various ways, making the semantic information of objects in the fused image more prominent. These approaches primarily rely on joint optimization with downstream tasks, injecting high-level semantic information into the fusion network via gradient backpropagation. While these methods can effectively inject semantic information to some extent, they all depend on guidance from external models or complex network designs.

In addition, image fusion faces another significant problem. Most fusion algorithms are not sufficiently robust for complex scenes. Specifically, if the input images are too dark (or overexposed) or have color casts, these methods often carry these defects directly into the final result. Consequently, some teams have designed specialized fusion networks to handle fusion problems in low-light environments~\cite{divfusion,ZVEFusion,IAIFNet}. However, because these methods are only adapted for low-light conditions, they tend to exhibit a degree of overexposure in normal lighting environments, indicating a lack of model flexibility. Recently, other methods~\cite{textif,omnifuse,promptfusion,Instruction-driven-fusion,controlfusion} have begun to explore whether user instructions can be used to more flexibly control the output of the fused image to adapt to different inputs (such as low light or color casts). Due to the powerful representation and generation capabilities of the underlying model, these methods have achieved stunning results. However, compared with the single mode image editing task, it is very challenging to achieve fine-grained and flexible control in image fusion. This is because it requires the model to understand the high-level semantics embedded in user instructions, while performing low-level alignment and integration of multimodal image content. Existing methods struggle to balance these two aspects, as control signals (such as those from text) cannot be effectively injected into these methods, thereby limiting their impact on actual fusion dynamics.
Moreover, many existing approaches depend on powerful external networks, such as pretrained instruction-following models or generative models, to interpret user intent and modify image content. As a result, they often fail to integrate visual and linguistic information within a unified latent space, which limits their ability to perform truly instruction-aware fusion in an end-to-end manner.

Meanwhile, if we want to achieve end-to-end controllable fusion within a single model, large-scale datasets are required for model training. The volume of data in existing image fusion datasets is too small, making it difficult to support the training for such a task. Therefore, the question is whether it is possible to develop a method that can both possess a strong capability for understanding high-level semantic information and achieve end-to-end controllable fusion within a single model, all while working with the limited scale of existing fusion data.

\begin{figure}[tb] 
  \centering 
  \includegraphics[width=\columnwidth]{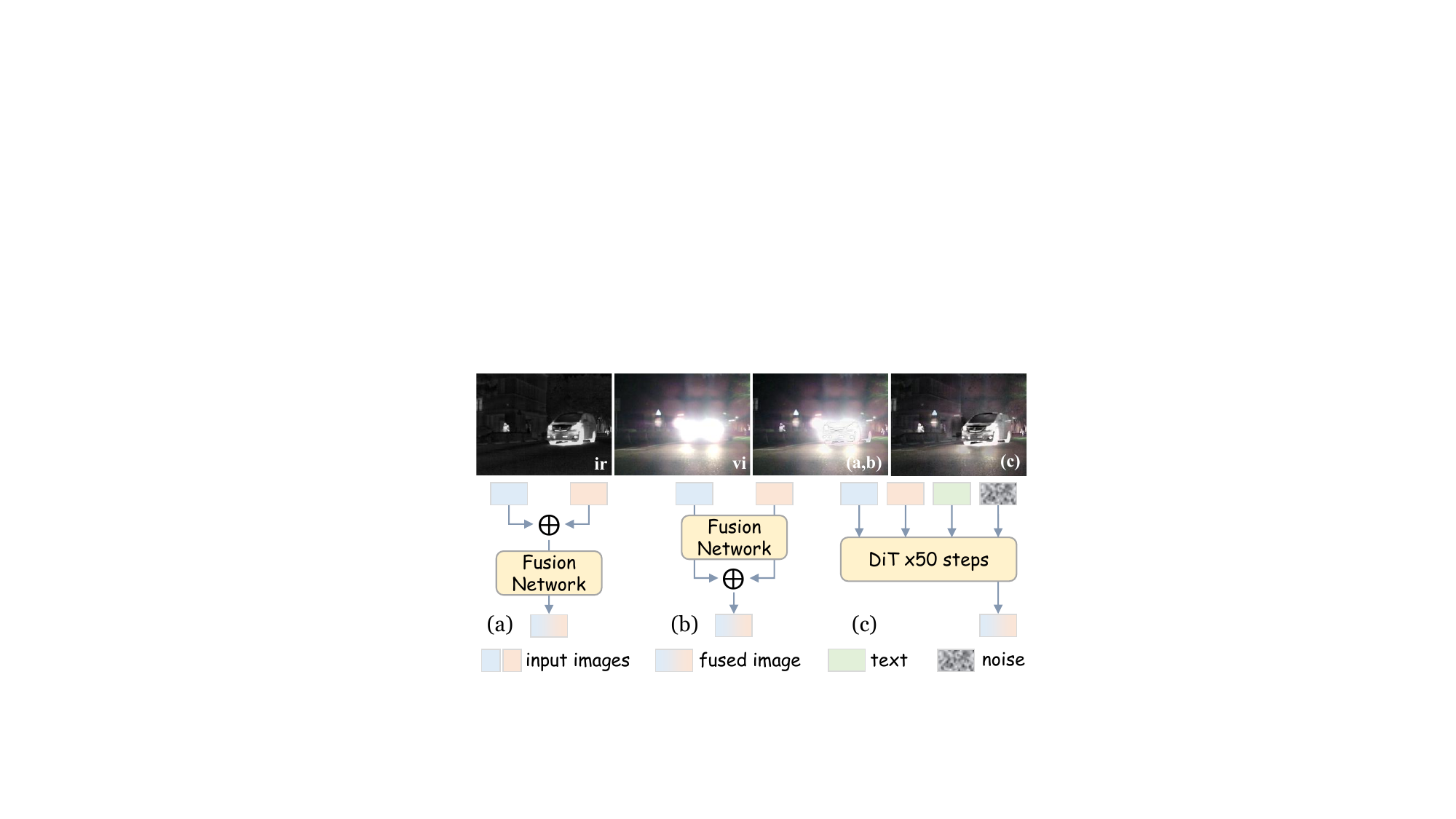} 
  \caption{(a) corresponds to pre-fusion. (b) corresponds to post-fusion. (c) represents our DiT-based method; The images in the first row display the fusion results of two different architectures under over-exposure conditions.}
  \label{fig:intro_net}
  \vspace{-18pt}
\end{figure}

To address these challenges, we propose DiTFuse, a novel method for end-to-end controllable fusion within a single model that imbues the final results with high-level semantic information. Unlike existing fusion networks that mostly adopt pre-fusion~\cite{fusiongan,ddcgan,seafusion} or post-fusion~\cite{swinfusion,vfusion,maefuse} models, as illustrated in Fig.~\ref{fig:intro_net}~(a,b), DiTFuse employs a pre-trained Diffusion Transformer (DiT~\cite{dit}) model as its backbone. Unlike conventional architectures, DiT enables joint modeling of vision and language through iterative denoising in a shared latent space, making it particularly suitable for injecting semantic control into the fusion process from early to late stages. It takes text and two images as parallel inputs, enabling the text to directly control the selection and generation of fused content. This provides stronger control compared to existing methods, which first obtain fused features and then apply textual control.

\begin{figure}[t] 
  \centering 
  \includegraphics[width=\columnwidth]{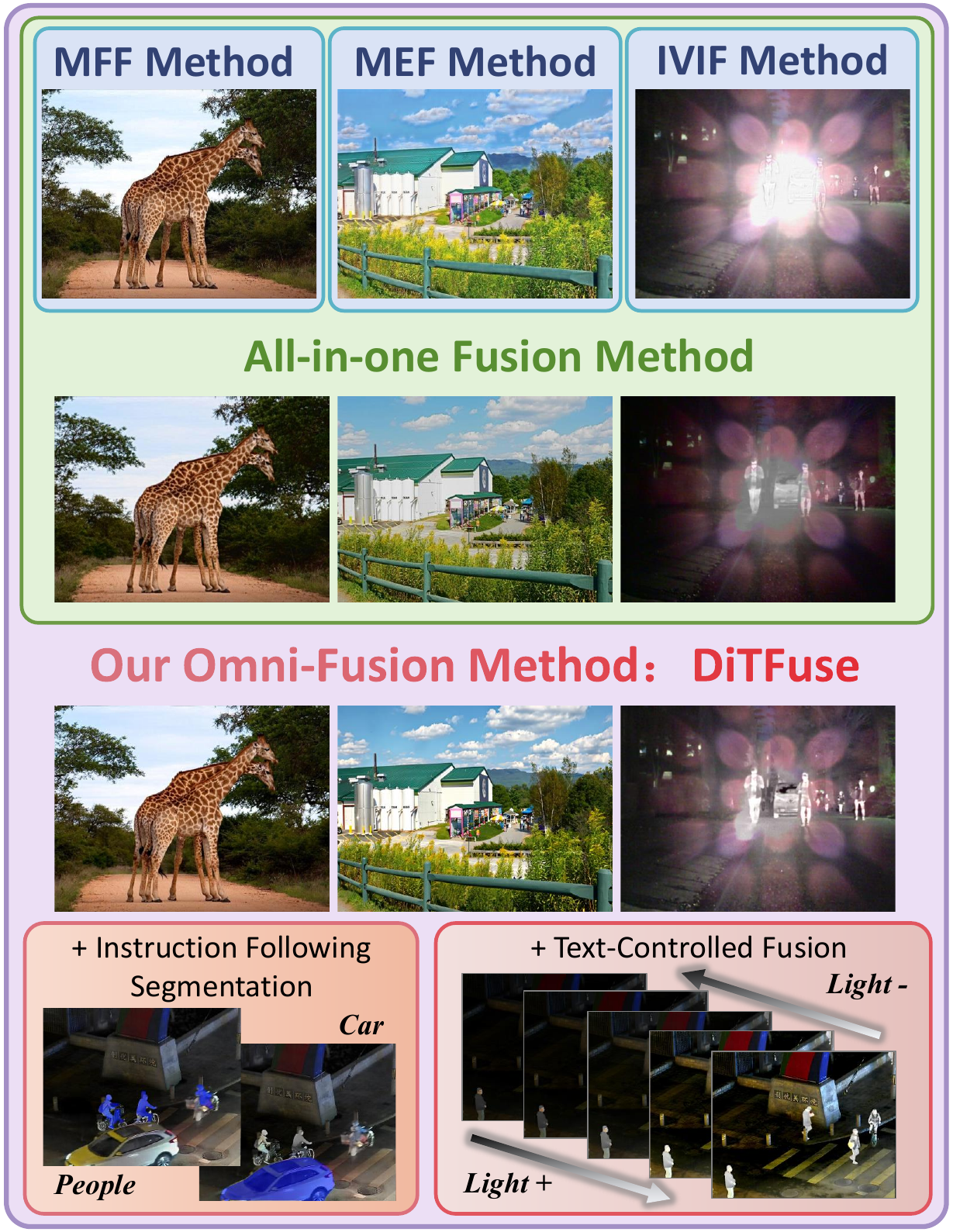} 
  \captionsetup{skip=2.5pt}
  \caption{Comparison of fusion method capabilities. Task-specific methods (e.g., MFFGAN~\cite{zhang2021mff}, CRMEF~\cite{liu2023embracing}, SeAFusion~\cite{seafusion}) handle only one fusion type. All-in-one methods (e.g., U2Fusion~\cite{u2fusion}) unify tasks but lack interactivity. Our method, DiTFuse, not only supports unified omni-fusion across MFF, MEF, and IVIF, but also enables text-guided fusion and instruction-following segmentation.}
  \label{fig:intro_re}
  \vspace{-18pt}
\end{figure}

Furthermore, the limited scale of existing fusion datasets makes it difficult to effectively fine-tune DiT, and the lack of ground truth for fusion tasks makes constructing large-scale training sets even more challenging. To address this, we innovatively propose M3 (\textbf{M}ulti-Degrade-\textbf{M}ask-Image-\textbf{M}odeling): we generate a vast number of complementary image pairs by applying random degradations (such as noise, blur, and masking) to large-scale natural images. Since the base DiT model has not been trained on fusion-like tasks, it cannot align the two input images at the pixel level, causing the output to not adhere well to the originals. These complementary images help the model learn to align the texture and structure of the two images. Moreover, the complementary degradation simulates a scenario where the two source images have different perceptual qualities and information content in the same region, which in turn prompts the model to select and fuse the parts with richer information. Based on this foundation, we constructed a mixed training paradigm using four types of data: fusion, M3, segmentation, and control. Within this, the training data from the segmentation task enables the model to learn high-level semantic information, thereby endowing it with stronger semantic perception capabilities during fusion tasks. Meanwhile, the M3 data, combined with a very small set of real fusion pairs, allows the model to learn various fusion modes (e.g., IVIF, MFF, MEF). 

In addition, to realize truly WYSIWYG controllable fusion, we constructed a large-scale, text-guided, multi-task dataset of control pairs based on M3 (see Fig.~\ref{fig:data_pipline} in the Methods section for details). This enables a complete end-to-end, instruction-driven fusion pipeline where the output content can be both controlled globally and adjusted hierarchically, shown in Fig.~\ref{fig:intro_re}. Finally, because we utilize data from segmentation tasks to guide the model in learning high-level semantic information, our model achieves, for the first time, a unification of fused image generation and downstream fusion tasks.

The main contributions of this work can be summarized as follows:
\begin{itemize}

\item \textbf{Multimodal parallel architecture for image fusion.} We propose a DiT-based framework (using Phi-3) that adopts a parallel input structure to fuse text and features for multi-modal fusion (e.g., IVIF, MFF, MEF). This design establishes a robust foundation to address modal redundancy in pre- and post-fusion stages.


\item \textbf{Multi-objective hybrid self-supervised training skill.} The self-supervised framework has three key elements: M3 generates complementary noisy image pairs from natural data to preserve realistic image priors for pixel-level synthesis; infrared-visible mean fusion brides modality gaps; while multi-prompt text-conditioned data ensures global control. This enables scalable multimodal alignment without ground-truth data, overcoming training sample limitations.



\item \textbf{Instruction-driven end-to-end controllable fusion.} We present the first end-to-end framework capable of directly controlling fusion results through text-based instructions, achieved by aligning text and multi-modal features within the latent space. This enables precise control over the visual effects of the fusion as well as the generation of instruction-following segmentation results from the multi-modal input. DiTFuse delivers accurate and tunable fusion effects and exhibits remarkable zero-shot capabilities on related tasks.


\end{itemize}

The remainder of this paper is organized as follows. Section~\ref{sec2} reviews deep learning-based fusion methods and the multimodal foundation models utilized in this paper. Section~\ref{sec3} provides detailed descriptions of the various design details of our work and the data construction pipline. Section~\ref{sec4} presents both quantitative and qualitative analysis results of DiTFuse on public datasets, along with ablation studies and extended analyses of our method's performance. Finally, Section~\ref{sec5} summarizes our approach and offers a forward-looking perspective.

\section{Related Work}\label{sec2}
In this section, we review the related work, which we divide into two primary categories. The first covers deep learning-based image fusion methods pertinent to our approach. The second provides an overview of the multimodal foundation models utilized in this paper.
\subsection{Deep Learning-based Image Fusion}

\subsubsection{Image Fusion Network Architectures}
The evolution of image fusion networks begin with pure CNN architectures~\cite{emfusion,multi,meta_learning,in3,searching,densefuse}, which utilized multi-scale convolutional blocks to extract and fuse local features during the upsampling stage. To enhance the generation of fine-grained details, these CNN backbones were later augmented with GAN branches~\cite{fusiongan,ddcgan,attentionfgan}. More recently, Transformer modules have been incorporated to capture long-range dependencies via self-attention, leading to sophisticated hybrid frameworks like CDDFuse~\cite{cddfuse}. This model integrates a CNN-based local branch with a Transformer-based global branch to process both texture and contextual information within a single network.

Despite these architectural advancements, a common limitation persists. Most existing methods still rely on simple pre-fusion~\cite{fusiongan,ddcgan,seafusion} or post-fusion~\cite{maefuse,cddfuse,omnifuse,s4fusion,li2025conti} schemes. This reliance on what is essentially information superposition restricts the model's ability to flexibly and dynamically adjust the weights and contributions of each modality during the fusion process.

Moreover, conventional methods often fail to properly retain high-level semantic information from the input images. To address this semantic degradation, subsequent works began exploring task-driven optimization. Initial efforts involved training a task-driven fusion network for segmentation~\cite{seafusion,liu2023multi} and detection~\cite{tardal,DetFusion,metafusion} after obtaining the fusion results. Subsequently, PSFusion~\cite{PSFusion} integrated fusion and segmentation networks and jointly trained the encoder, enhancing the semantic awareness of the fused results. Recently, some methods~\cite{wu2025every,liu2025dcevo} have injected downstream-task semantic information directly into the latent space, thereby enhancing the model’s ability to perceive and preserve semantics. These methods demonstrated improved semantic retention in challenging environments. However, all of these approaches depend on guidance from an additional downstream-task network. Our method requires no extra architectural components: it can perform segmentation directly, enabling the model to acquire high-level semantic information naturally in the process.

Concurrently, current research is exploring language-guided fusion to generate user-customizable outputs. For instance, Text-IF~\cite{textif} leverages CLIP~\cite{clip} embeddings with AdaIN~\cite{adain} modulation for text-adaptive fusion, while TeRF~\cite{terf} combines LLaMA~\cite{llama} with GroundingDINO-SAM~\cite{grounding} to enable region-specific control. Nevertheless, these methods adhere to a \textit{``fuse-then-edit"} paradigm. This involves fusing content at the feature or image level, and subsequently controlling the output at that level. This approach inherently decouples the fusion and control processes, rather than achieving a direct mapping from linguistic commands to the fusion operation itself.

\subsubsection{All-In-One Image Fusion Methods}
In parallel, significant research has focused on developing a single network architecture that can flexibly handle diverse fusion types, such as Infrared-Visible Image Fusion (IVIF), Multi-Focus Fusion (MFF), and Multi-Exposure Fusion (MEF). However, these approaches~\cite{ifcnn,pmgi,u2fusion,swinfusion,defusion,liang2024fusion} typically employed an identical network and training process for all functions.

Given that the optimization objectives for these tasks are inherently distinct, this one-size-fits-all strategy leads to compromised overall fusion performance. To address this limitation, we introduce a task token to partially decouple the three tasks. This mechanism prevents them from being forced along a single, unified optimization trajectory, thereby enabling more effective, task-specific adjustments.

\subsubsection{MIM Adaptations for Fusion Tasks}
The absence of ground-truth makes Masked Image Modeling (MIM) particularly valuable for self-supervised fusion training. DeFusion~\cite{defusion} pioneered this paradigm by generating complementary masks to pretrain fusion models, explicitly guiding networks to recover both shared and modality-specific features through partial image reconstruction. DDBFusion~\cite{ddbfusion} enhanced this framework through adaptive masking strategies incorporating blur and illumination variations, which strengthened complementary feature learning. While these methods demonstrate promising results in homogeneous fusion tasks (e.g., multi-focus fusion), their requirements of precise separation of shared and unique components face inherent challenges in cross-modal scenarios like infrared and visible image fusion. Specifically, the significant modality gap complicates explicit decomposition of ``common" versus ``unique" features, limiting the applicability of standard MIM strategies to heterogeneous fusion tasks.

\subsubsection{Diffusion-based Image Fusion Methods}
The emergence of diffusion models has introduced new possibilities for image fusion. DDFM~\cite{ddfm} was the first to apply Diffusion models for image fusion by aligning fused outputs with both infrared and visible inputs, though its pixel-wise similarity objectives inherently tends toward mean fusion. Later, Diff-IF~\cite{diffif} attempted to address this limitation by introducing fusion priors into the denoising process. However, this method relies on existing fusion results as guidance, thereby creating a circular dependency that fundamentally constrains the model's performance ceiling. Recent work like E2E-MFD~\cite{e2e} introduced task-aware conditioning into the diffusion process to enhance downstream applicability. However, these methods structurally treat multi-modal inputs by simply overlaying them in the output, failing to achieve true information selection and fusion.


\subsection{Unified Multimodal Models}

The advent of large-scale models has spurred significant research into unifying image understanding and generation. The central hypothesis is that a model's deep comprehension of content can powerfully inform and guide the process of image synthesis.

Pioneering efforts in this domain, such as Chameleon~\cite{chameleon}, approached this by first discretizing images into tokens using a VQ-GAN~\cite{vqgan}. Subsequently, they synthesized images by autoregressively predicting each token in sequence. However, this autoregressive (AR) methodology has an inherent limitation: it is prone to accumulating and propagating errors, which causes the fidelity of later image tokens to degrade. To overcome the shortcomings of AR, subsequent works evolved their strategy. Show-O~\cite{show-o}, for example, retained the discrete VQ-GAN codes but pivoted to a parallel masked-token generation approach inspired by MaskGIT~\cite{maskgit}. This allowed the model to generate all image tokens simultaneously, mitigating the issue of error accumulation. Recognizing that the discretization process itself still sacrifices fine-grained details, a further evolution led to models like Omnigen~\cite{omnigen} and Transfusion~\cite{transfusion}, which transitioned to operating in a continuous latent space. These frameworks employ a DiT-style diffusion backbone to denoise these continuous latents, a method that has proven to yield exceptionally high-quality images while maintaining robust text-based semantic understanding.

Since our work is focused exclusively on the image generation aspect and does not require text output, we adopt Omnigen as our base model, leveraging its proven capabilities for high-fidelity image synthesis.

\section{Method}\label{sec3}


This section first details our DiT-based fusion architecture, which unifies two input images and a textual instruction within a single latent diffusion framework. We explain how multimodal tokens are constructed, organized, and interacted through a hybrid attention scheme to realize controllable, semantics-aware fusion. Subsequently, we present the training strategy: a flow-matching objective optimized under a multi-task setting that spans classical fusion, image-level control, segmentation, and our proposed Multi-Degrade-Mask-Image-Modeling (M3). Finally, we describe the construction pipeline of the training data, including how complementary degradations, task-aware targets, and standardized prompts are generated to scale supervision while maintaining consistency across tasks.


\begin{figure*}[t] 
  \centering 
  \includegraphics[width=\textwidth]{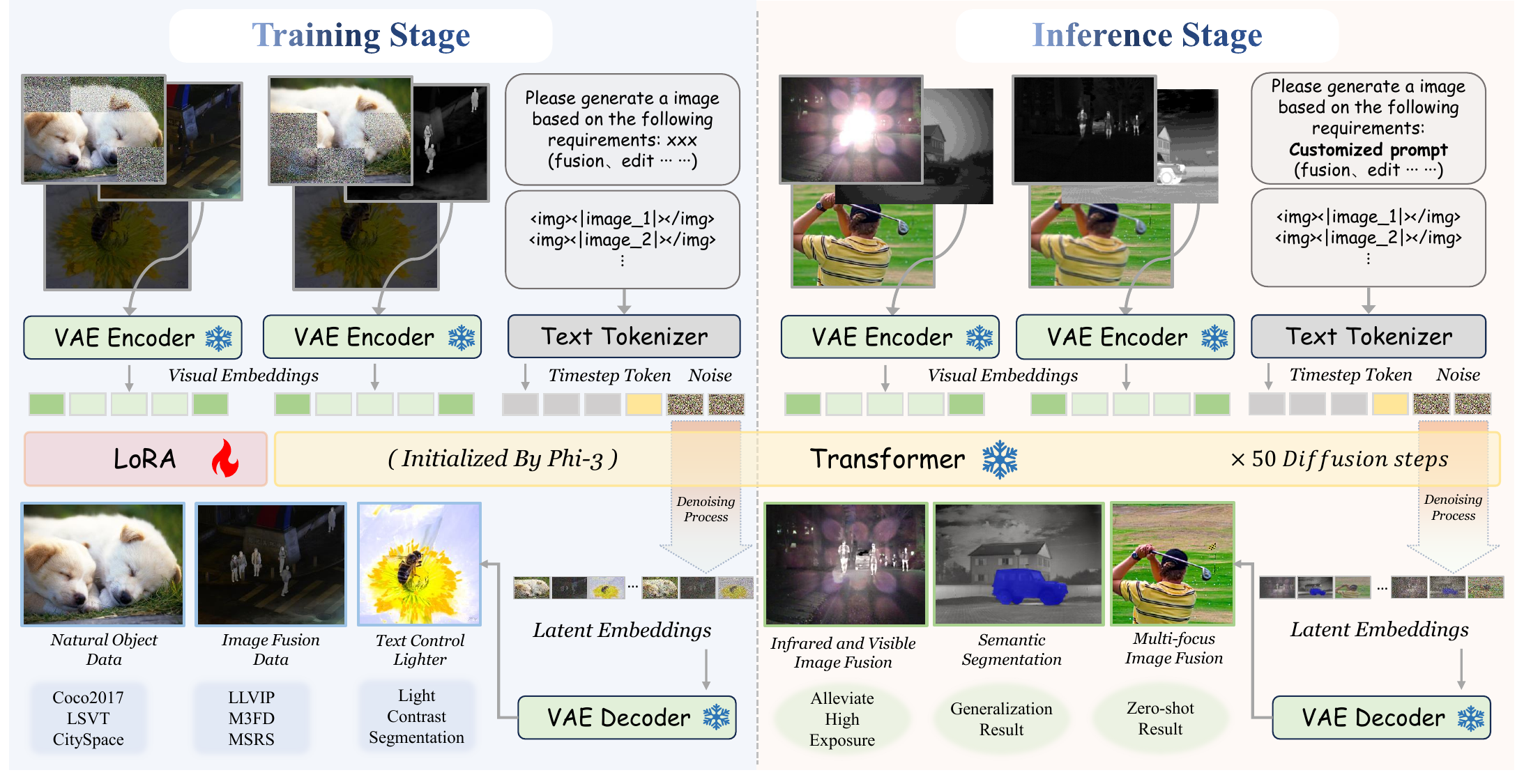} 
  \captionsetup{skip=1.5pt}
  \caption{The framework of our model. Textual control information is encoded through the Text Tokenizer, and image information is encoded into Visual Embeddings via VAE. These are used together as conditional information to control the denoising process of DiT. The left half of the diagram represents the training stage, while the right half represents the inference stage. During the training stage, we primarily use the M3 method to constrain the model, and during the inference stage, it can work on multiple fusion tasks.}
  \vspace{-5pt}
  \label{fig:net}
\end{figure*}

\subsection{Model Design}

We build our image fusion model upon the standard Diffusion Transformer (DiT) architecture~\cite{dit}, which we then instruction-fine-tune starting from the pre-trained Omnigen~\cite{omnigen} model. A key challenge lies in how to effectively encode and integrate multimodal information into a unified latent space for joint generation. In this section, we elaborate on the overall design from three perspectives: model structure, input representation, and attention mechanism.

\subsubsection{Model Structure}

Our model follows the DiT-based diffusion generation paradigm, with additional components to support multimodal inputs and task conditioning. As illustrated in Fig.~\ref{fig:net}, the architecture comprises four main modules:

\begin{itemize}
    \item \textbf{Text Encoder}: We utilize the pretrained Phi-3 tokenizer~\cite{phi3} to convert natural language instructions into semantic embeddings. Given an instruction $\mathcal{T}$, the text encoder $\text{Transformer}_{\text{enc}}$ generates contextualized tokens $\mathbf{T} \in \mathbb{R}^{L \times d_t}$ using a multi-head transformer:
    \begin{equation}
        \mathbf{T} = \text{Transformer}_{\text{enc}}(\text{Tokenize}(\mathcal{T})),
    \label{eq:token}
    \end{equation}
    where $L$ is the token length and $d_t=3072$ is the embedding dimension.

    \item \textbf{Visual Encoder}: We adopt the VAE module from SDXL~\cite{sdxll} as our image encoder $\text{VAE}_{\text{enc}}$. Each input image $\mathbf{X} \in \mathbb{R}^{H \times W \times 3}$ is first patchified and then encoded into latent tokens $\mathbf{V} \in \mathbb{R}^{\frac{H}{8} \times \frac{W}{8} \times C}$:
    \begin{equation}
        \mathbf{V} = \text{VAE}_{\text{enc}}(\text{Patchify}_{8\times8}(\mathbf{X})).
        \label{eq:vae}
    \end{equation}

    \item \textbf{DiT Blocks}: Following the DiT design, we concatenate all image and text tokens into a single input sequence, which is processed by a stack of $N=32$ transformer layers to model cross-modal interactions and generate fused outputs.

    \item \textbf{LoRA Adaptation}: To adapt the pretrained DiT model to fusion-specific tasks while retaining its multimodal capabilities, we apply Low-Rank Adaptation (LoRA)~\cite{lora} to all linear layers in the transformer blocks. For a weight matrix $\mathbf{W} \in \mathbb{R}^{d_{in} \times d_{out}}$, the adapted version is:
    \begin{equation}
        \mathbf{W}' = \mathbf{W} + \alpha \cdot \mathbf{B}\mathbf{A}^\top, \quad \mathbf{B} \in \mathbb{R}^{d_{in} \times r}, \mathbf{A} \in \mathbb{R}^{d_{out} \times r},
    \end{equation}
    where the rank $r=64$ and scaling factor $\alpha=0.5$.
\end{itemize}

\subsubsection{Input Representation}

The model accepts both visual and textual inputs. Initially, we utilize the Phi-3 tokenizer defined in Eq.~\ref{eq:token} to handle the text. For the image, we begin by employing the VAE in Eq.~\ref{eq:vae} along with a basic linear layer to obtain the latent representation. Next, we apply linear embedding to each patch in the latent space, transforming it into a sequence of visual tokens. The patch size is set to 2, and standard frequency-based position embeddings are used for the visual tokens. Before integrating the image sequences into the text token sequence, each image sequence is encapsulated using the special tokens \texttt{<img>} and \texttt{</img>}. Additionally, a timestep embedding is appended at the end of the input sequence to enhance temporal representation. 

Through the aforementioned encoding scheme, all inputs are converted into a unified representation and fed in parallel into the DiT model for interpretation. This design facilitates stronger cross-modal interaction, enabling the fusion model to jointly and fully consider the visual information in the input image and the control instructions expressed in text. The combined information constrains the denoising process of image generation. Specifically, by virtue of the Transformer architecture, computing the text queries (Q) against the image keys (K) helps highlight salient visual cues while down-weighting less relevant content, thereby achieving text-driven content selection and generation. Compared with traditional approaches, this scheme markedly enhances instruction comprehension and controllability.

\begin{figure*}[t] 
  \centering 
  \includegraphics[width=\textwidth]{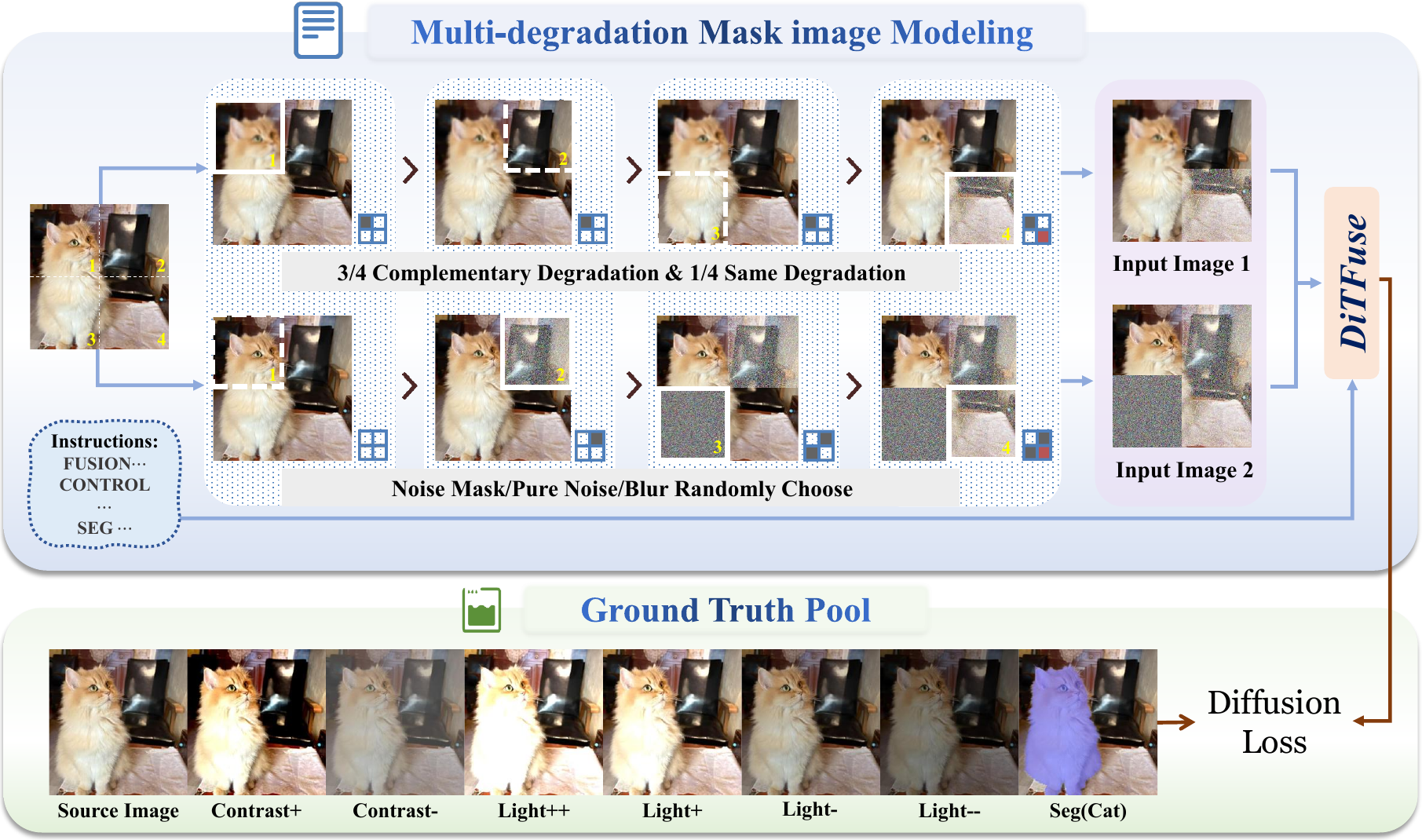} 
  \captionsetup{skip=1.5pt}
  \caption{Training data construction pipeline.
The blue box illustrates the input images generation process, which follows the Multi-degradation Mask image Modeling (M3) strategy. The orange box presents the Ground Truths corresponding to different instructions. For M3 training data, the Ground Truth is the source image itself. For control-type instructions, the Ground Truth is created by adjusting the image’s overall contrast or brightness. For segmentation instructions, the Ground Truth is generated by overlaying a blue transparent mask on the target class.}
  \vspace{-5pt}
  \label{fig:data_pipline}
\end{figure*}

\begin{figure}[h!] 
  \centering 
  \includegraphics[width=\columnwidth]{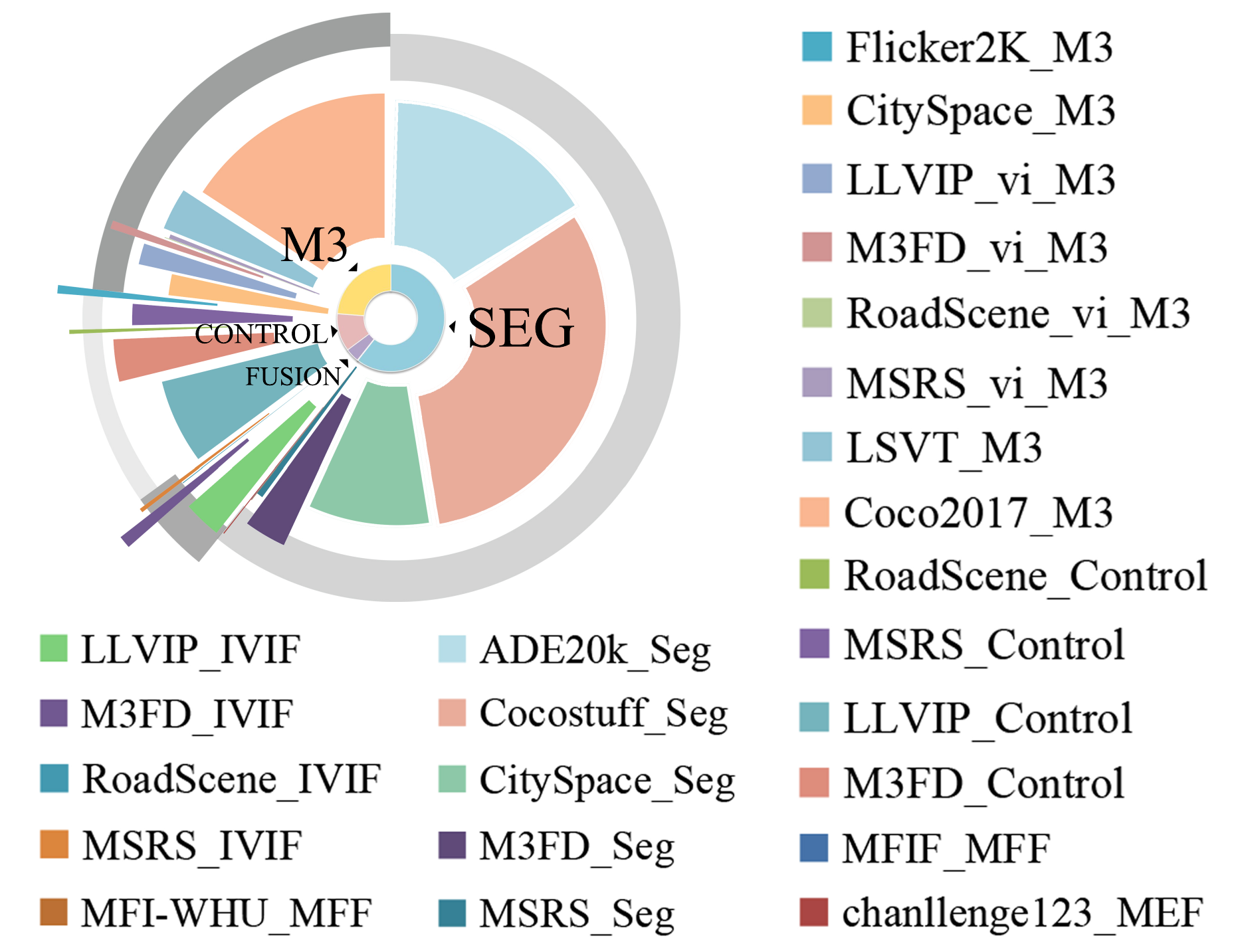} 
  \captionsetup{skip=1pt}
  \caption{The diagram is a sunburst chart showing the data composition. The inner ring displays the four main data categories, and the outer ring shows the proportion of each specific dataset within those categories.}
  \vspace{-5pt}
  \label{fig:data_component}
\end{figure}

\subsubsection{Attention Mechanism}

During the learning process, images require global information awareness and, therefore, cannot utilize causal attention in the same manner as textual data. To address this, we apply causal attention to each element within the sequence while employing bidirectional attention within individual image sequences. This design enables each patch to attend to other patches within the same image while ensuring that each image can only attend to preceding images or text sequences in the overall sequence.

This hybrid attention strategy not only allows text instructions to interact with all visual tokens for fine-grained semantic control, but also preserves the structural integrity of individual images through localized bidirectional attention. By restricting inter-image attention to a causal form, the model respects the sequential dependency between modalities, enabling progressive fusion guided by instruction-aware context. This facilitates more coherent, interpretable, and controllable fusion results, especially in scenarios involving multi-source or instruction-conditioned fusion tasks.

\subsection{Training Strategy}

In this section, we will primarily cover two aspects: the training mechanism and the training data construction pipeline.

\subsubsection{Training Mechanism}

In the field of image fusion, a central challenge is the lack of reliable ground truth in real-world scenarios. This difficulty is further compounded by the subtly different optimization objectives across fusion tasks: multi-exposure fusion aims for exposure balancing; infrared–visible fusion focuses on preserving rich semantic content; and multi-focus fusion emphasizes spatial sharpness across regions. Such task-specific objectives make it hard to define a universal supervision signal. Moreover, collecting high-quality fusion data is often costly and labor-intensive, which severely constrains the scalability of supervised training for large models. Consequently, fusion-specific supervision alone provides insufficient guidance for learning generalizable representations. In addition, for models based on the DiT architecture, achieving pixel-level alignment between the two input images for fusion is itself a challenging problem.

To mitigate this limitation, we adopt a multi-task training strategy that leverages a diverse set of auxiliary tasks designed to strengthen the model’s complementary capabilities. As illustrated in Fig.~\ref{fig:data_component}, our training set consists of four categories of samples: segmentation (Seg), multi-degradation mask image modeling (M3), image-level visual control (Control), and classical fusion data. Notably, only 4.3\% of the training data—13,692 out of 316,392 samples—comes from fusion tasks. The majority of the data is contributed by M3 and segmentation tasks, which can be constructed at scale and offer meaningful supervision from alternative perspectives.

The M3 data is particularly effective at strengthening the model’s ability to integrate multi-source degraded inputs and achieve accurate reconstruction. In M3, two complementary input images are constructed via masks, and the model is trained to output a complete image. The critical step is aligning the boundaries of the masked image blocks so that the result remains continuous and complementary, thereby enabling the model to learn pixel-level alignment. In parallel, segmentation data enhances the model’s semantic understanding, helping it recognize object categories and more faithfully align visual content with textual instructions.

All tasks are unified under a single instruction-based framework: the model receives two images and a textual instruction describing the desired operation, and is trained to produce a task-consistent output. We adopt a flow-matching based learning mechanism, which differs from traditional denoising diffusion models by replacing the iterative denoising process with a continuous interpolation between clean data and noise.

Formally, for a given target image $\mathbf{x}$ and Gaussian noise $\boldsymbol{\epsilon} \sim \mathcal{N}(0,1)$, we sample a timestep $t \in [0, 1]$ and construct a noised version as:
\begin{equation}
    \mathbf{x}_t = t\mathbf{x} + (1 - t)\boldsymbol{\epsilon}.
\end{equation}
The model takes as input the tuple $(\mathbf{x}_t, t, c)$, where $c$ denotes the conditioning instruction, and predicts the target velocity. The training objective is to minimize the mean squared error between the ground truth and the predicted velocity:
\begin{equation}
    \mathcal{L} = \mathbb{E} \left[ \left\| (\mathbf{x} - \boldsymbol{\epsilon}) - v_\theta(\mathbf{x}_t, t, c) \right\|^2 \right].
\end{equation}
This unified training objective enables joint optimization across diverse tasks—including fusion, control, and segmentation—under a single learning framework. By sharing supervision signals across tasks and training objectives, the model develops robust and generalizable capabilities while maintaining a simplified pipeline. This also marks the first time that various types of fusion tasks and downstream fusion tasks have been unified within a single architecture.

\subsubsection{Training Data Construction Pipeline}

The construction of both input and supervision in our training data follows a task-aware but unified procedure. As shown in Fig.~\ref{fig:data_pipline}, the M3 samples are built by introducing multi degradations over a source image. To do this, we first duplicate the image into two views and partition each into grids of $16 \times 16$, $32 \times 32$, or $64 \times 64$ patches. Then, for a large majority of patches—typically 75\%—we apply complementary degradation: in each image, only one of the two patches at a given location is degraded, using randomly chosen operations from adding noise masks, Gaussian noise, or blur. The remaining 25\% of patches are jointly degraded in both views, enforcing a shared corruption across inputs.

This mixture encourages the model to reason over complementary information and learn to recover the most reliable regions. The complementary degradation helps it develop a preference for well-preserved areas from noisy sources, while the shared degradation pushes it to reconstruct clean signals even when no clear reference exists. This data construction approach simulates the selection and fusion of the most salient and clear information from two input images. In doing so, it helps promote the training for various fusion tasks while mitigating the key issue of data scarcity.

The ground truth varies depending on the type of instruction provided. When the instruction corresponds to M3, the model is trained to reconstruct the original source image. For infrared and visible image fusion, we use the mean fusion results as the ground truth. For multi-exposure image fusion, we rely on existing synthetic datasets, where the normally exposed image is typically used as the target. For multi-focus image fusion, we also use public datasets, which usually provide a pseudo ground truth for supervision. In the control setting, we modify the contrast or brightness of the target image based on textual prompts, encouraging the model to adjust image-level properties in line with language guidance. For segmentation instructions, the supervision consists of softly masked images where the target category is highlighted with a translucent blue overlay.
\begin{figure}[t] 
  \centering 
  \includegraphics[width=\columnwidth]{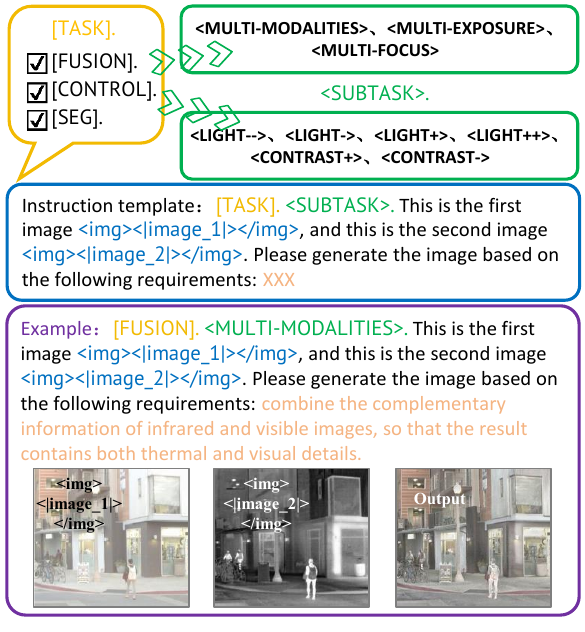} 
  \captionsetup{skip=1pt}
     \caption{Illustration of the prompt structure. Each prompt consists of four components: a \texttt{[TASK]} tag, a \texttt{<SUBTASK>} tag, two image placeholders \texttt{<img><image\_n></img>}, and a free-form textual instruction. For fusion tasks, the tag is \texttt{[FUSION]} with subtags \texttt{<MULTI-MODALITIES>}, \texttt{<MULTI-EXPOSURE>}, or \texttt{<MULTI-FOCUS>}. For image control tasks, the tag is \texttt{[CONTROL]} with one of the subtags \texttt{<LIGHT++>}, \texttt{<LIGHT+>}, \texttt{<LIGHT->}, \texttt{<LIGHT-->}, \texttt{<CONTRAST+>}, or \texttt{<CONTRAST->}. Segmentation tasks use the tag \texttt{[SEG]} without a subtask. For M3 data, only the general tag \texttt{[FUSION]} is used, without any subtask tag.}
    \vspace{-5pt}
    \label{fig:prompt_structure}
\end{figure}

To maintain consistency across different task types, each training sample is paired with a standardized prompt that combines an explicit task tag, a subtask tag (if applicable), image placeholders, and a free-form instruction. As illustrated in Fig.~\ref{fig:prompt_structure}, this structure provides both compositional flexibility and clarity. The task tags (\texttt{[FUSION]}, \texttt{[CONTROL]}, \texttt{[SEG]}) categorize the high-level objectives, while subtags such as \texttt{<MULTI-MODALITIES>}, \texttt{<LIGHT++>}, or others offer more granular guidance. This unified format enables the model to distinguish among task types and optimize accordingly, reducing potential ambiguity in multi-task training.

\begin{figure*}[h] 
  \centering 
  \includegraphics[width=\textwidth]{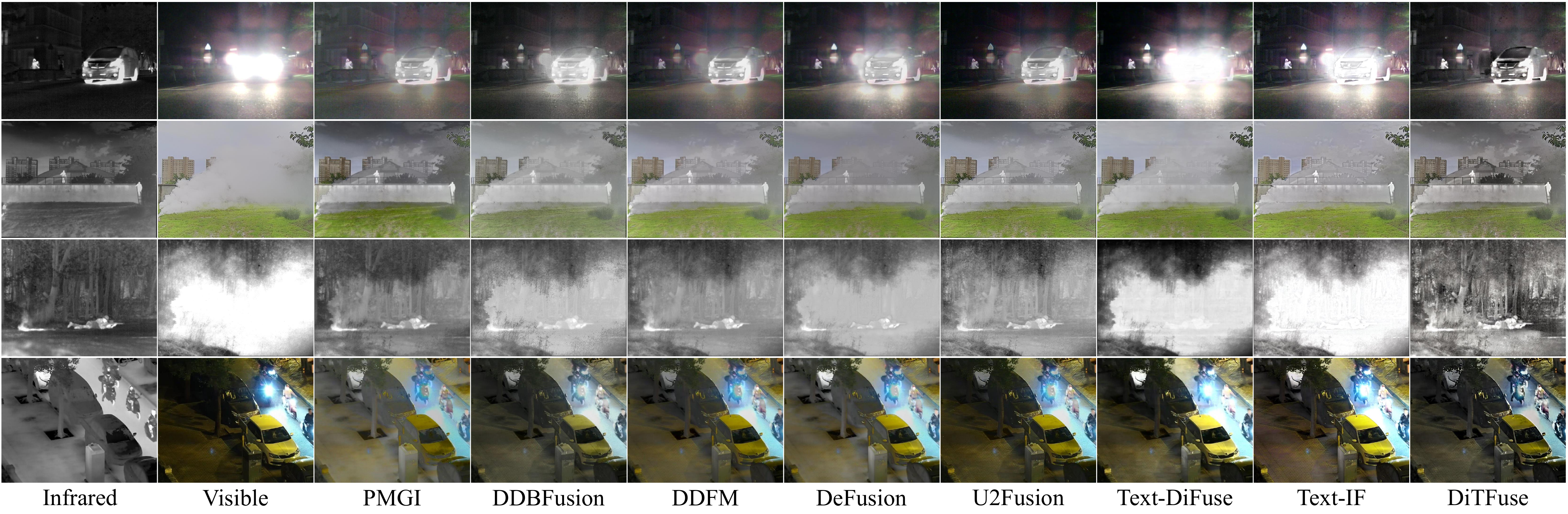} 
  \captionsetup{skip=1pt}
  \caption{Qualitative comparison of our DiTFuse (the base fusion prompt) and existing image fusion methods. From top to bottom: one group of data from MSRS, one group of data from $\text{M}^{3}\text{FD}$, one group of data from TNO, and one group of data from LLVIP datasets, respectively. Please zoom in for a better view.}
  \vspace{-5pt}
  \label{fig:ivif_cmp}
\end{figure*}


Overall, the design of our training data ensures that every sample, regardless of origin, is paired with a meaningful target and an interpretable instruction. This consistency across heterogeneous tasks allows us to scale training without introducing fragmentation or requiring task-specific loss engineering.

\section{Experiments}\label{sec4}

In this section, we first provide some implementation details of our experiment. Subsequently, we conduct comprehensive comparisons with state-of-the-art methods across several fusion datasets, evaluated from both qualitative and quantitative perspectives. Additionally, we demonstrate text-controllable fusion scenarios enabled by our framework. We then conduct adequate ablation studies to validate the effectiveness of our training strategy. Furthermore, we explore the zero-shot capability of the model.

\begin{table*}[h]
\centering
\caption{Quantitative comparison of our DiTFuse (the base fusion prompt) and existing image fusion methods on the MSRS, $\text{M}^{3}\text{FD}$, and TNO datasets. The best and second best results of each metric are highlighted in \textbf{bold} and \underline{underlined}, respectively.}
\normalsize
\setlength{\tabcolsep}{3pt} 
\begin{adjustbox}{width=1.00\textwidth,center}
\begin{tabular}{@{}l@{\hspace{2pt}}c@{\hspace{8pt}}*{15}{@{\hspace{2pt}}c@{\hspace{2pt}}}@{}}
\toprule
\multirow{2}{*}{\large Method} & \multirow{2}{*}{} & 
 \multicolumn{5}{c}{\large MSRS} & 
\multicolumn{5}{c}{\large $\text{M}^3\text{FD}$} & 
\multicolumn{5}{c}{\large TNO} \\
\cmidrule(lr){3-7} \cmidrule(lr){8-12} \cmidrule(lr){13-17}
&  & \hspace{-2pt} MSE$\downarrow$  & \hspace{-2pt} PSNR$\uparrow$  & MANIQA$\uparrow$  & LIQE$\uparrow$  & \hspace{-2pt} CLIPIQA+$\uparrow$ 
& MSE$\downarrow$  & \hspace{-2pt} PSNR$\uparrow$  & MANIQA$\uparrow$  & LIQE$\uparrow$ &   \hspace{-2pt} CLIPIQA+$\uparrow$ 
& MSE$\downarrow$  & \hspace{-2pt} PSNR$\uparrow$  & MANIQA$\uparrow$  & LIQE$\uparrow$ & \hspace{-2pt} CLIPIQA+$\uparrow$  \\
\midrule
SwinFusion   & \textcolor{gray}{JAS'22}  & \hspace{-2pt} 0.038 & \hspace{-2pt} 64.52 & 0.138 & 1.108 & 0.312 & 0.059 & \hspace{-2pt} 61.37 & 0.284 & \underline{1.704} & \hspace{-2pt} 0.481 & 0.059 & \hspace{-2pt} 61.36 & 0.164 & 1.010 & \hspace{-2pt} 0.229 \\
\addlinespace[0.5pt]
SeAFusion    & \textcolor{gray}{IF'22} & \hspace{-2pt} 0.038 & \hspace{-2pt} 64.33 & 0.144 & \underline{1.138} & 0.355 & 0.060 & \hspace{-2pt} 61.12 & 0.288 & 1.641 &  \hspace{-2pt} 0.466 & 0.058 & \hspace{-2pt} 61.63 & 0.187 & 1.013 & \hspace{-2pt} 0.267 \\
\addlinespace[0.5pt]
PMGI     & \textcolor{gray}{AAAI'20} & \hspace{-2pt} 0.066 & \hspace{-2pt} 60.32 & 0.142 & 1.030 & 0.244 & 0.038 & \hspace{-2pt} 62.91 & 0.277 & 1.359 & \hspace{-2pt} 0.420 & 0.044 & \hspace{-2pt} 62.23 & 0.162 & 1.013 & \hspace{-2pt} 0.212 \\
\addlinespace[0.5pt]
DDBFusion    & \textcolor{gray}{IF'25} & \hspace{-2pt} \textbf{0.021} & \hspace{-2pt} \textbf{66.93} & 0.138 & 1.102 & 0.285 & \textbf{0.032} & \hspace{-2pt} \underline{63.81} & 0.274 & 1.445 & \hspace{-2pt} 0.440 & 0.039 & \hspace{-2pt} 62.97 & 0.199 & 1.019 & \hspace{-2pt} 0.244 \\
\addlinespace[0.5pt]
DDFM         & \textcolor{gray}{ICCV'23} & \hspace{-2pt} \underline{0.022} & \hspace{-2pt} 66.60 & 0.142 & 1.053 & \underline{0.296} & \underline{0.033} & \hspace{-2pt} 63.58 & \underline{0.296} & 1.553 & \hspace{-2pt} 0.452 & 0.045 & \hspace{-2pt} 62.21 & 0.187 & 1.019 & \hspace{-2pt} 0.253 \\
\addlinespace[0.5pt]
DeFusion     & \textcolor{gray}{ECCV'22} & \hspace{-2pt} 0.026 & \hspace{-2pt} 66.08 & 0.132 & 1.042 & 0.318 & 0.036 & \hspace{-2pt} 63.52 & 0.276 & 1.425 & \hspace{-2pt} 0.433 & 0.040 & \hspace{-2pt} \underline{63.40} & 0.185 & 1.015 & \hspace{-2pt} 0.253 \\
\addlinespace[0.5pt]
U2Fusion      & \textcolor{gray}{PAMI'22} & \hspace{-2pt} \underline{0.022} & \hspace{-2pt} 66.46 & \underline{0.154} & 1.096 & 0.327 & 0.033 & \hspace{-2pt} 63.61 & 0.282 & 1.423 &  \hspace{-2pt} \textbf{0.506} & \underline{0.038} & \hspace{-2pt} 63.08 & \underline{0.201} & 1.014 & \hspace{-2pt} 0.269 \\
\addlinespace[0.5pt]
Text-DiFuse  & \textcolor{gray}{NIPS'24} &  \hspace{-2pt} 0.092 & \hspace{-2pt} 58.62 & 0.131 & 1.084 & 0.284 & 0.058 & \hspace{-2pt} 60.88 & 0.275 & 1.497 & \hspace{-2pt} 0.423 & 0.056 & \hspace{-2pt} 61.19 & 0.197 & \textbf{1.027} & \hspace{-2pt} 0.270 \\
\addlinespace[0.5pt]
Text-IF      & \textcolor{gray}{CVPR'24}  & \hspace{-2pt} 0.039 & \hspace{-2pt} 64.10 & 0.140 & 1.107 & \underline{0.362} & 0.051 & \hspace{-2pt} 62.10 & 0.286 & 1.661 & \hspace{-2pt} 0.457 & 0.051 & \hspace{-2pt} 62.02 & 0.190 & \underline{1.023} & \hspace{-2pt} \underline{0.281} \\
\addlinespace[0.5pt]
\textbf{DiTFuse} &    & \hspace{-2pt} \textbf{0.021} & \hspace{-2pt} \underline{66.63} & \textbf{0.162} & \textbf{1.240} & \textbf{0.392} & \textbf{0.032} & \hspace{-2pt} \textbf{63.81} & \textbf{0.299} & \textbf{1.718} & \hspace{-2pt} \underline{0.498} & \textbf{0.036} & \hspace{-2pt} \textbf{63.50} & \textbf{0.209} & 1.019 & \hspace{-2pt} \textbf{0.297} \\
\bottomrule
\end{tabular}
\end{adjustbox}

\label{tab:ivif_tab}

\begin{minipage}{\textwidth}
\footnotesize
\centering
\end{minipage}
\end{table*}

\subsection{Implementation Details}

The proposed method is evaluated on eight datasets, which include five for infrared-visible image fusion (MSRS~\cite{piafusion}, $\text{M}^{3}\text{FD}$~\cite{tardal}, TNO~\cite{TNO}, RoadScene~\cite{roadscene}, and LLVIP~\cite{llvip}), two for multi-focus image fusion (MFF~\cite{mfif} and RealMFF~\cite{real_mff}), and one for multi-exposure image fusion (SICE~\cite{sice}). The model is trained for 2 epochs on 8×A100 GPUs with a global batch size of 64, using a learning rate of $1 \times 10^{-4}$ and single-step gradient accumulation. We incorporate LoRA layers (rank=64) for efficient parameter tuning and set the conditional dropout probability to 0.01 for modality robustness.

\subsection{Evaluation of Infrared-Visible Image Fusion}

We conduct both qualitative and quantitative analyses to compare 9 different fusion approaches, including SwinFusion~\cite{swinfusion}, SeAFusion~\cite{seafusion}, PMGI~\cite{pmgi}, DDBFusion~\cite{ddbfusion}, DDFM~\cite{ddfm}, DeFusion~\cite{defusion}, U2Fusion~\cite{u2fusion}, Text-DiFuse~\cite{Text-DiFuse}, and Text-IF~\cite{textif}. Among them, PMGI, SwinFusion, DDBFusion, U2Fusion, and DeFusion, like our method, are all-in-one models capable of performing Infrared-Visible Image Fusion (IVIF), Multi-Focus Image Fusion (MFF), and Multi-Exposure Fusion (MEF) tasks. To be more specific, SwinFusion is a traditional method that uses gradient loss for guidance. SeAFusion is a fusion method driven by downstream tasks. DDBFusion and DeFusion are trained using the Masked Image Modeling (MIM) strategy. DDFM and Text-DiFuse are fusion methods based on diffusion models, while both Text-DiFuse and Text-IF explore text-controlled fusion.

\textbf{Qualitative comparison.} We select three primary scene types for comparison: over-exposure scenes, foggy scenes, and low-light scenes, as illustrated in Fig.~\ref{fig:ivif_cmp}. The over-exposure scenes correspond to the first and second rows. Compared to other methods, our approach more effectively removes over-exposure artifacts while enhancing the visibility of pedestrians and vehicles. The foggy scenes are shown in the third and fourth rows. Our method demonstrates superior performance in fog removal and in highlighting human figures and architectural structures, while also enhancing the texture details from the infrared image. The low-light scene corresponds to the final row. By utilizing a pre-trained model, our generated results are visually more consistent with natural image priors, producing clearer images with fewer infrared artifacts. To a certain extent, it also effectively highlights information in regions with poor initial brightness. Overall, our method effectively preserves both semantic and texture information in a visually perceptible manner.

\begin{figure*}[t] 
  \centering 
  \includegraphics[width=\textwidth]{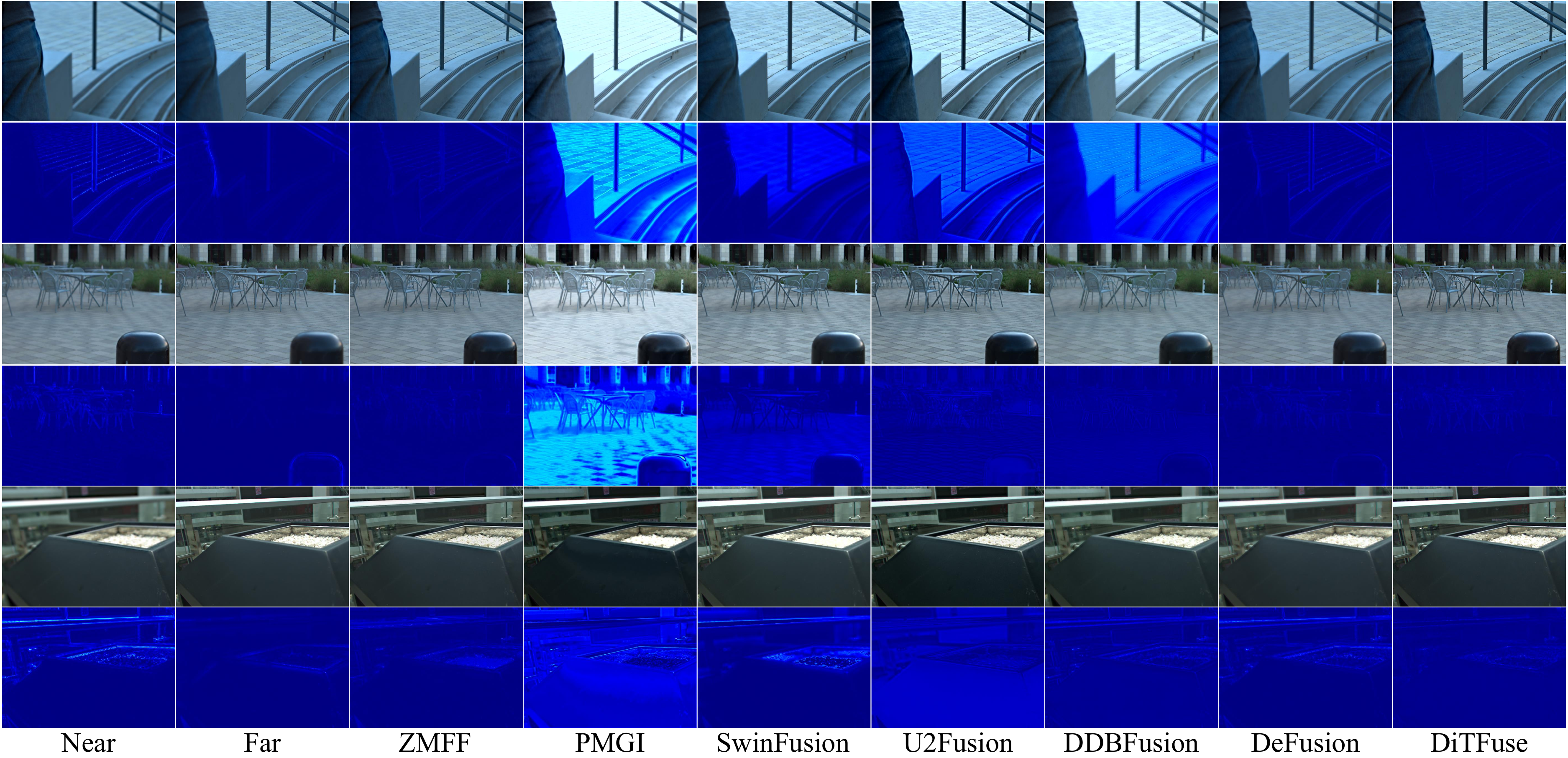} 
  \caption{Qualitative comparison of DiTFuse with six MFF methods on three multi-focus image pairs on the RealMFF dataset. We provide the residual maps for each result of comparison and input images to highlight the difference with GT. Please zoom in for a better view.}
  \vspace{-5pt}
  \label{fig:mfif_cmp}
\end{figure*}

\textbf{Quantitative comparison.} Our method has achieved SOTA performance on the MSRS, $\text{M}^{3}\text{FD}$, and TNO datasets across multiple benchmark evaluations. The quantitative results are presented in Tab.~\ref{tab:ivif_tab}. Among these, MSE and PSNR are reference-based metrics that measure the consistency of pixel-level information between the fused image and the source images. These results indicate that our method effectively preserves texture information from both modalities, which is a key advantage of our proposed M3 training strategy. Meanwhile, MANIQA, LIQE, and CLIPIQA+ are no-reference metrics. They assess not only the texture fidelity of the generated image at multiple scales but also the clarity and completeness of its semantic information. These metrics show that our generated images are neither overly smoothed nor deficient in critical semantic information. In summary, these complementary evaluation results jointly validate the robustness of our method in balancing low-level visual fidelity with high-level semantic perception.

\subsection{Evaluation of Multi-Focus Image Fusion}

We conduct both qualitative and quantitative analyses to compare 6 different fusion approaches, including ZMFF~\cite{ZMFF}, PMGI~\cite{pmgi}, SwinFusion~\cite{swinfusion}, U2Fusion~\cite{u2fusion}, DDBFusion~\cite{ddbfusion}, and DeFusion~\cite{defusion}. Among them, PMGI, SwinFusion, DDBFusion, U2Fusion, and DeFusion, like our method, are all-in-one models, whereas ZMFF is a method that specializes in the multi-focus image fusion (MFF) task.

\textbf{Qualitative comparison.} For our comparison, we choose three pairs of images from the RealMFF dataset, as shown in Fig.~\ref{fig:mfif_cmp}. In contrast to other all-in-one methods, our approach more effectively fuses the near- and far-focus images. The residual maps reveal that our method successfully integrates texture details from both focal planes, thereby enriching the final fused image. This success stems from our M3 training strategy, which empowers the model to select the clearest regions from the input images. Moreover, even when compared against ZMFF—a specialized MFIF method—our approach exhibits a superior capacity for information fusion.

\begingroup
\setlength{\tabcolsep}{2pt}  
\renewcommand{\arraystretch}{1.0}  
\begin{table*}[h]
  \centering
  \caption{Quantitative comparison of our DiTFuse (the base fusion prompt) and existing image fusion methods on the MFIF, RealMFF, and SICE datasets. MFIF and RealMFF are multi-focus datasets, while SICE is a multi-exposure dataset. The best and second best results of each metric are highlighted in \textbf{bold} and \underline{underlined}, respectively.}
  \normalsize        
  \begin{adjustbox}{max width=\textwidth}
  \begin{tabular}{lcccccccccclccccc}
    \toprule
    \multirow{2}{*}{\large{Method}} &
    \multicolumn{5}{c}{\large{MFIF (MFF\_DATA)}} &
    \multicolumn{5}{c}{\large{RealMFF (MFF\_DATA)}} &
    \multirow{2}{*}{\large{Method}} &
    \multicolumn{5}{c}{\large{SICE (MEF\_DATA)}} \\
    \cmidrule(lr){2-6}
    \cmidrule(lr){7-11}
    \cmidrule(lr){13-17}
    & SF$\uparrow$ & AG$\uparrow$ & LIQE$\uparrow$ & MUSIQ$\uparrow$ & CLIPIQA+$\uparrow$ 
    & SF$\uparrow$ & AG$\uparrow$ & LIQE$\uparrow$ & MUSIQ$\uparrow$ & CLIPIQA+$\uparrow$ 
    && EN$\uparrow$ & SD$\uparrow$ & LIQE$\uparrow$ & MUSIQ$\uparrow$ & CLIPIQA+$\uparrow$ \\
    \midrule
    ZMFF & 16.77 & 5.863 & 2.476 & 49.32 & 0.483 & 14.86 & 5.703 & \underline{3.037} & 53.23 & \underline{0.566} 
            & HSDS\_MEF & 7.286 & 0.268 & \underline{3.774} & 69.30 & 0.667 \\
    PMGI & 11.36 & 4.226 & 2.195 & 53.63 & 0.581 & 15.43 & 5.884  & 2.610 & 52.55 & 0.511  
            & PMGI & 6.677 & 0.156 & 2.592 & \textbf{70.72} & 0.599 \\
    SwinFusion & 20.16 & 6.941 & 2.994 & 58.53 & 0.628 & \textbf{18.32} & \underline{6.577} & 2.943 & \underline{55.72} & 0.543
            & SwinFusion & 7.133 & 0.271 & 3.571 & 67.50 & 0.665 \\
    U2Fusion & \underline{22.11} & \underline{8.060} & \underline{3.119} & \underline{62.14} & \underline{0.658} & 16.64 & 6.370 & 2.021 & 53.40 & 0.501
            & U2Fusion & 7.328 & 0.262 & 3.346 & 67.67 & 0.673 \\
    DDBFusion & 15.51 & 5.374  & 2.795 & 57.33 & 0.623 & 14.19 & 5.302 & 2.683 & 53.31 & 0.534
            & DDBFusion & \underline{7.453} & \underline{0.268} & 3.684 & 69.30 & \underline{0.686} \\
    DeFusion & 12.60 & 4.679 & 2.856 & 58.95 & 0.619 & 12.41 & 4.700 & 2.421 & 51.81 & 0.517
            & DeFusion & 7.315 & 0.240 & 3.567 & 68.97 & 0.650 \\
    DiTFuse & \textbf{23.81} & \textbf{8.260} & \textbf{3.891} & \textbf{68.42} & \textbf{0.668} & \underline{18.26} & \textbf{6.634} & \textbf{3.408} & \textbf{58.46} & \textbf{0.572}  
            & DiTFuse & \textbf{7.532} & \textbf{0.274} & \textbf{4.005} & \underline{70.32} & \textbf{0.693} \\
    \bottomrule
  \end{tabular}
  \end{adjustbox}
  \label{tab:mff_mef_tab}
\end{table*}
\endgroup

\begin{figure*}[h] 
  \centering 
  \includegraphics[width=\textwidth]{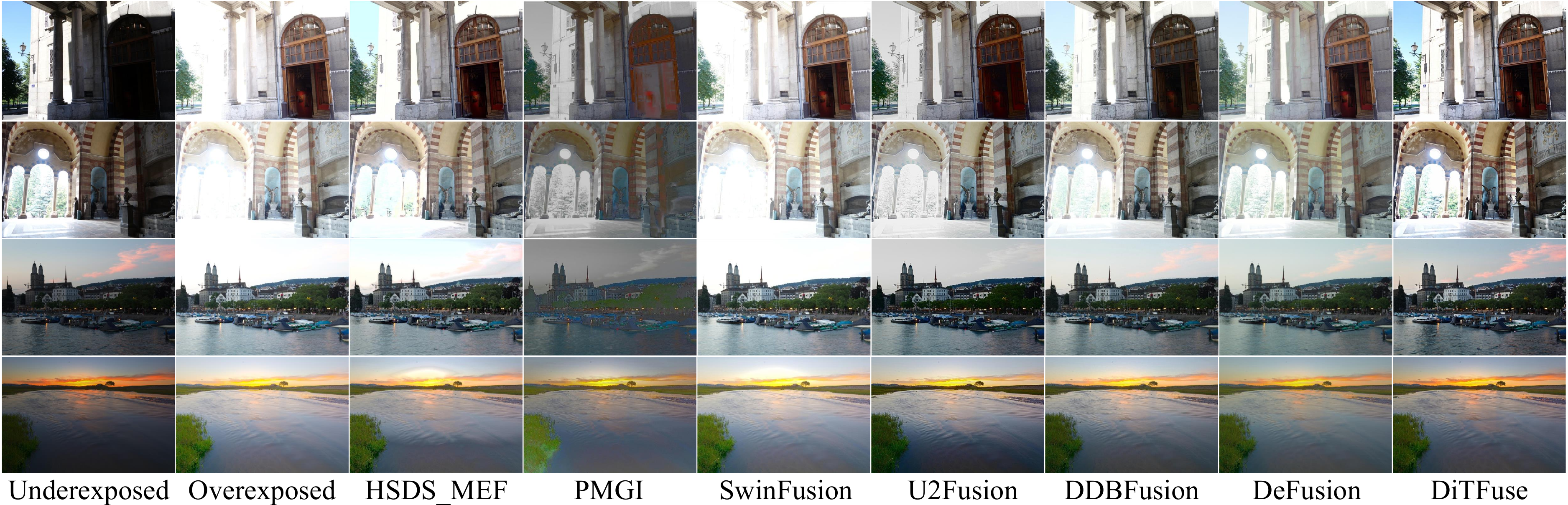} 
  \captionsetup{skip=1pt}
  \caption{Qualitative comparison of our DiTFuse (the base fusion prompt) and existing image fusion methods. All four groups of data are from SICE dataset. Please zoom in for a better view.}
  \vspace{-5pt}
  \label{fig:mef_cmp}
\end{figure*}

\begin{figure*}[h] 
  \centering 
  \includegraphics[width=\textwidth]{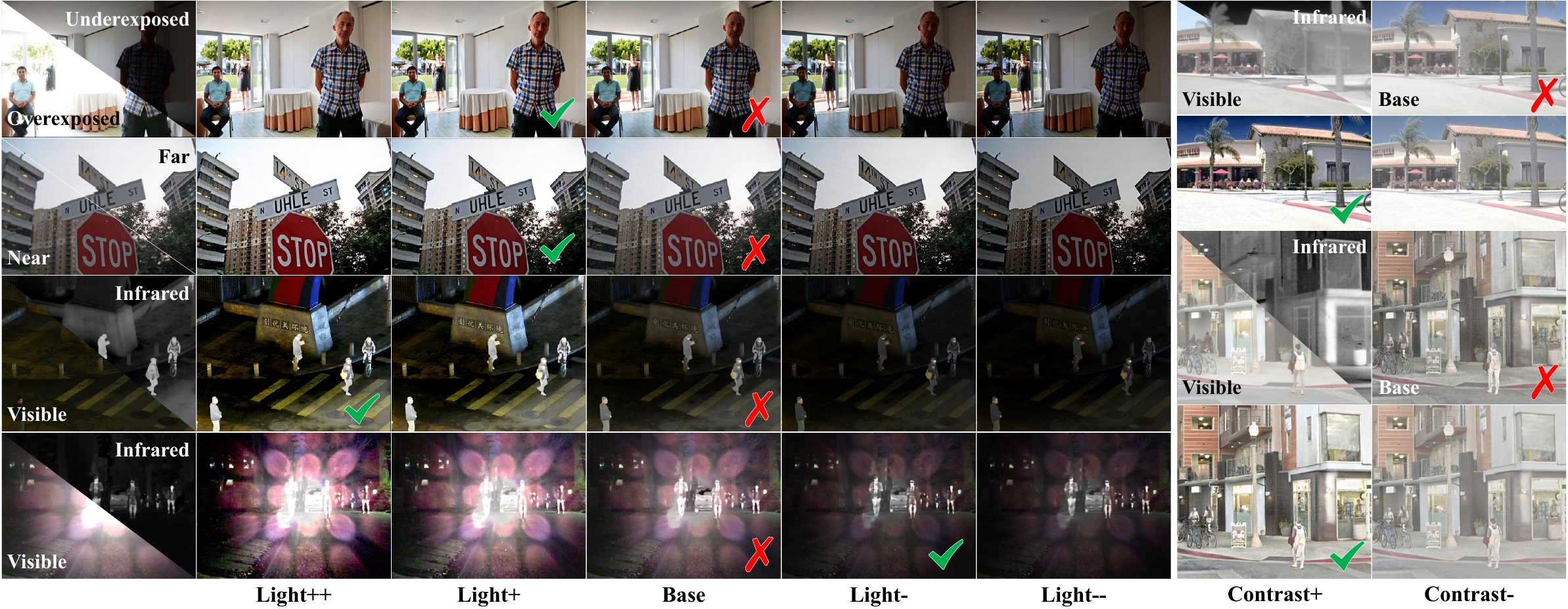} 
  \captionsetup{skip=1pt}
  \caption{The images from the SICE, MFIF, LLVIP, MSRS, TNO, and RoadScene datasets utilize prompt control to adjust the brightness and contrast of the fused image.}
  \vspace{-5pt}
  \label{fig:eg_control}
\end{figure*}

\begin{figure*}[h] 
  \centering 
  \includegraphics[width=\textwidth]{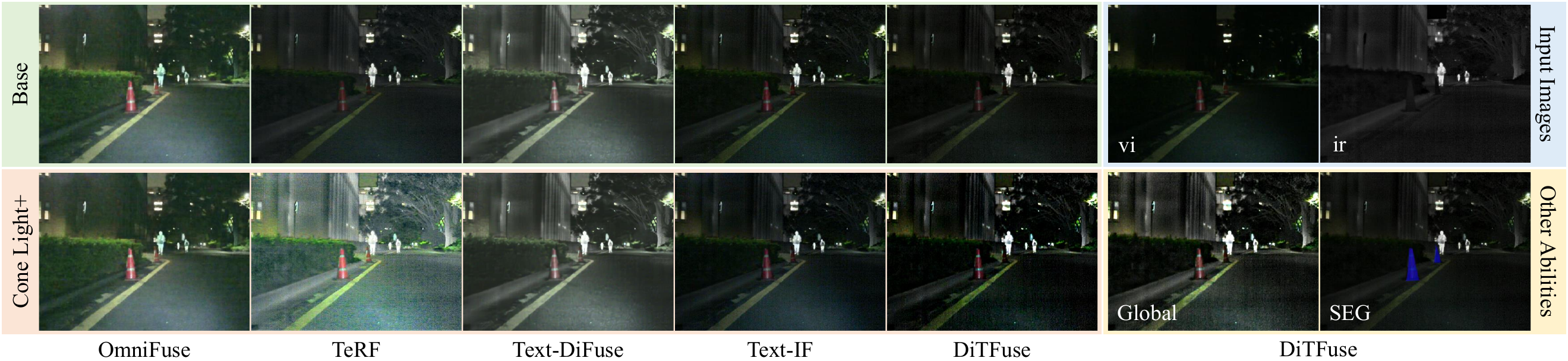} 
  \captionsetup{skip=1pt}
  \caption{Qualitative comparison of text-controlled fusion methods. The first row shows each method’s base fusion results; the second row shows the results after increasing the brightness for `cone’.}
  \vspace{-5pt}
  \label{fig:eg_control_cmp}
\end{figure*}

\textbf{Quantitative comparison.} Our method has achieved SOTA performance on the MFIF and RealMFF datasets across multiple benchmark evaluations. The quantitative experimental results are presented in Tab.~\ref{tab:mff_mef_tab}. Among these, SF (Spatial Frequency) and AG (Average Gradient) respectively measure the texture complexity and gradient richness of the generated image. These results indicate that our method can better select regions with more prominent textures and clearer information. Meanwhile, LIQE, MUSIQ, and CLIPIQA+ are no-reference metrics. They assess not only the texture fidelity of the generated image at multiple scales but also the clarity and completeness of its semantic information. In summary, these complementary evaluation results jointly validate that our method can better integrate information to produce clearer and sharper multi-focus fused images.

\subsection{Evaluation of Multi-Exposure Image Fusion}

We conduct both qualitative and quantitative analyses to compare six different fusion approaches, including HSDS\_MEF~\cite{HSDS_mef}, PMGI~\cite{pmgi}, SwinFusion~\cite{swinfusion}, U2Fusion~\cite{u2fusion}, DDBFusion~\cite{ddbfusion}, and DeFusion~\cite{defusion}. Among them, PMGI, SwinFusion, DDBFusion, U2Fusion, and DeFusion, like our method, are all-in-one models, whereas HSDS\_MEF is a method that specializes in the multi-exposure image fusion (MEF) task. The data for our fusion process are selected from the SICE dataset, specifically using the pairs of images with exposure values of -1 EV and +1 EV.

\textbf{Qualitative comparison.} For our comparison, we select four pairs of images from the SICE dataset, as depicted in Fig.~\ref{fig:mef_cmp}. Our multi-exposure fusion results are more effective at preserving color and detail information. In overexposed scenes, unlike other methods that tend to produce averaged-out results, our approach better mitigates the effects of overexposure by selecting the portions from the underexposed input image that are not affected by it.

\textbf{Quantitative comparison.} Our method achieves SOTA performance on the SICE dataset across multiple benchmark evaluations. The quantitative experimental results are presented in Tab.~\ref{tab:mff_mef_tab}. For our evaluation, we adopt a suite of metrics including EN, SD, LIQE, MUSIQ, and CLIPIQA+. This set of metrics provides a comprehensive assessment from both pixel-level and semantic viewpoints, gauging the information richness of our fusion results and confirming that the images we produce are both natural and clear.

\subsection{Evaluation of Text Control Capability}

We showcase several results generated under user guidance, demonstrating multi-level control over brightness and contrast, as depicted in Fig.~\ref{fig:eg_control}. This controllability proves highly effective across a diverse range of fusion tasks. In this context, `base' denotes a standard, unguided fusion result. Since this base output often falls short of the desired outcome, we provide users with an intuitive, WYSIWYG control mechanism that effectively overcomes the limitations of a fixed fusion model. For instance, brightness control enables users to modify the image’s luminance via specific prompts, thereby elevating the visual quality of the final fusion. This feature is especially valuable in low-light and multi-exposure scenarios, where it helps mitigate the loss of semantic detail caused by poor illumination. Our DiTFuse architecture, which decouples input processing from output generation, facilitates a more dynamic integration of information from both modalities. This design circumvents the challenge of precisely controlling the fusion balance, a common problem in traditional pre-fusion and post-fusion methods that rely on direct image blending. Contrast control, meanwhile, allows for the enhancement or reduction of image contrast, lending a greater sense of structure and depth to the visual information. In situations where the source images are inherently monotonous in color, we leverage contrast enhancement to boost the chromatic richness of the fused output.

To demonstrate the superiority of the control ability of our model, we present a qualitative comparison with existing text-based control methods, as shown in Fig.~\ref{fig:eg_control_cmp}. In the scenario where brightness is increased only for the `cone’, our results achieve finer-grained control over the object region, while other methods fail to precisely target specific areas. In addition, our method supports global brightness control and can also produce segmentation results, as illustrated in Fig.~\ref{fig:eg_control_cmp}. These findings highlight the flexibility and diversity of control offered by our approach.


\begin{table}[t]
  \centering
  \caption{mIoU segmentation results comparing our method with LISA. Here, LISA(Fusion), LISA(VIS), and LISA(IR) denote segmentation performed on the fused image, the visible image, and the infrared image, respectively.}
  \resizebox{\columnwidth}{!}{
  \begin{tabular}{lcccc}
    \toprule
    \textbf{Class} & \textbf{DiTFuse} & \textbf{LISA(Fusion)} & \textbf{LISA(VIS)} & \textbf{LISA(IR)} \\
    \midrule
    Building    & 0.5271 & 0.5598 & \textbf{0.5654} & 0.5622 \\
    Bus         & \textbf{0.4087} & 0.2871 & 0.3142 & 0.2687 \\
    Car         & \textbf{0.6426} & 0.5829 & 0.5955 & 0.4621 \\
    Motorcycle  & \textbf{0.2903} & 0.1923 & 0.2790 & 0.0877 \\
    Person      & \textbf{0.3776} & 0.2238 & 0.2117 & 0.2684 \\
    Pole        & \textbf{0.2405} & 0.2193 & 0.2289 & 0.1184 \\
    Road        & 0.7436 & \textbf{0.7744} & 0.7640 & 0.7660 \\
    Sidewalk    & 0.2039 & 0.3143 & \textbf{0.3357} & 0.2673 \\
    Sky         & 0.8915 & 0.8869 & \textbf{0.8998} & 0.8674 \\
    Truck       & \textbf{0.3161} & 0.2695 & 0.2825 & 0.2037 \\
    Vegetation  & \textbf{0.5936} & 0.5137 & 0.5624 & 0.4659 \\
    \textbf{Overall} & \textbf{0.4760} & 0.4386 & 0.4581 & 0.3943 \\
    \bottomrule
  \end{tabular}}
  \vspace{3pt}
  \vspace{-5pt}
  \label{tab:ditfuse_lisa_per_class}
\end{table}

\begin{table*}[t]
\centering
\caption{Statistical results of the correctness ratios for the three dimensions of the GPT-4o evaluation, comparing the segmentation performance of LISA and DiTFuse. The values on the left are for LISA, and the values on the right are for DiTFuse. \textcolor{gray}{Gray} highlighting indicates the lower result.}
\small
\setlength{\tabcolsep}{5pt} 
\begin{adjustbox}{width=1.00\textwidth,center}
\begin{tabular}{@{}l@{\hspace{2pt}}*{12}{@{\hspace{2pt}}c@{\hspace{2pt}}}@{}}
\toprule
\multirow{2}{*}{\small Label}  & 
\multicolumn{3}{c}{\small $\text{M}^3\text{FD}$} & 
\multicolumn{3}{c}{\small RoadScene} & 
\multicolumn{3}{c}{\small SICE} & 
\multicolumn{3}{c}{\small RealMFF} \\
\cmidrule(lr){2-4} \cmidrule(lr){5-7} \cmidrule(lr){8-10} \cmidrule(lr){11-13} 
&    P.Ratio  & R.Ratio  & I.Ratio  &  P.Ratio  & R.Ratio  & C.Ratio  &  P.Ratio  & R.Ratio  & I.Ratio &  P.Ratio  & R.Ratio  & I.Ratio  \\
\midrule
Sky       & \textcolor{gray}{0.87}/0.94 & \textcolor{gray}{0.98}/0.99 & \textcolor{gray}{0.98}/0.99 & 0.84/\textcolor{gray}{0.80} & \textcolor{gray}{0.95}/0.98 & \textcolor{gray}{0.97}/0.99 & \textcolor{gray}{0.75}/0.84 & \textcolor{gray}{0.97}/0.98 & \textcolor{gray}{0.97}/0.98 & \textcolor{gray}{0.46}/0.71 & \textcolor{gray}{0.90}/0.93 & 0.94/\textcolor{gray}{0.92} \\
\addlinespace[0.5pt]
Tree      & 0.70/\textcolor{gray}{0.64} & \textcolor{gray}{0.76}/0.78 & \textcolor{gray}{0.86}/0.91 & \textcolor{gray}{0.49}/0.58 & \textcolor{gray}{0.42}/0.94 & \textcolor{gray}{0.81}/0.94 & 0.64/\textcolor{gray}{0.59} & \textcolor{gray}{0.79}/0.90 & \textcolor{gray}{0.90}/0.95 & 0.49/\textcolor{gray}{0.47} & 0.88/\textcolor{gray}{0.85} & \textcolor{gray}{0.86}/0.88 \\
\addlinespace[0.5pt]
Car        & \textcolor{gray}{0.66}/0.70 & \textcolor{gray}{0.76}/0.89 & \textcolor{gray}{0.98}/0.99 & \textcolor{gray}{0.67}/0.68 & \textcolor{gray}{0.59}/0.91 & \textcolor{gray}{0.94}/0.97 & \textcolor{gray}{0.18}/0.43 & \textcolor{gray}{0.86}/0.96 & \textcolor{gray}{0.97}/1.00 & \textcolor{gray}{0.34}/0.35 & \textcolor{gray}{0.50}/0.67 & 0.92/\textcolor{gray}{0.83} \\
\addlinespace[0.5pt]
Road        & \textcolor{gray}{0.25}/0.84 & \textcolor{gray}{0.78}/0.96 & 1.00/\textcolor{gray}{0.99} & \textcolor{gray}{0.24}/0.78 & \textcolor{gray}{0.50}/0.87 & \textcolor{gray}{0.33}/1.00  & \textcolor{gray}{0.14}/0.69 & \textcolor{gray}{0.67}/0.86 & \textcolor{gray}{0.40}/0.98 & \textcolor{gray}{0.22}/0.68 & \textcolor{gray}{0.75}/0.96 & \textcolor{gray}{0.67}/0.93 \\
\addlinespace[0.5pt]
Buildings   & \textcolor{gray}{0.69}/0.76 & 0.98/\textcolor{gray}{0.79} & \textcolor{gray}{0.98}/0.99 & 0.63/\textcolor{gray}{0.52} & 0.95/\textcolor{gray}{0.88} & \textcolor{gray}{0.91}/1.00 & \textcolor{gray}{0.64}/0.65 & 0.98/\textcolor{gray}{0.90} & \textcolor{gray}{0.96}/0.97 & \textcolor{gray}{0.65}/0.76 & 0.98/\textcolor{gray}{0.94} & \textcolor{gray}{0.97}/0.98 \\
\addlinespace[0.5pt]
Person      & \textcolor{gray}{0.51}/0.79 & \textcolor{gray}{0.63}/0.75 & \textcolor{gray}{0.94}/0.98 & \textcolor{gray}{0.43}/0.46 & \textcolor{gray}{0.49}/0.71 & 0.93/\textcolor{gray}{0.91} & \textcolor{gray}{0.12}/0.43 & \textcolor{gray}{0.84}/0.92 & 1.00/\textcolor{gray}{0.97} & \textcolor{gray}{0.30}/0.50 & \textcolor{gray}{0.88}/0.90 & \textcolor{gray}{0.96}/0.97 \\
\addlinespace[0.5pt]
\bottomrule
\end{tabular}
\end{adjustbox}
\vspace{-5pt}
\label{tab:seg_cmp}

\begin{minipage}{\textwidth}
\footnotesize
\centering
\end{minipage}
\end{table*}

\subsection{Evaluation of Text-Controlled Segmentation Capability}

\begin{figure}[t] 
  \centering 
  \includegraphics[width=\columnwidth]{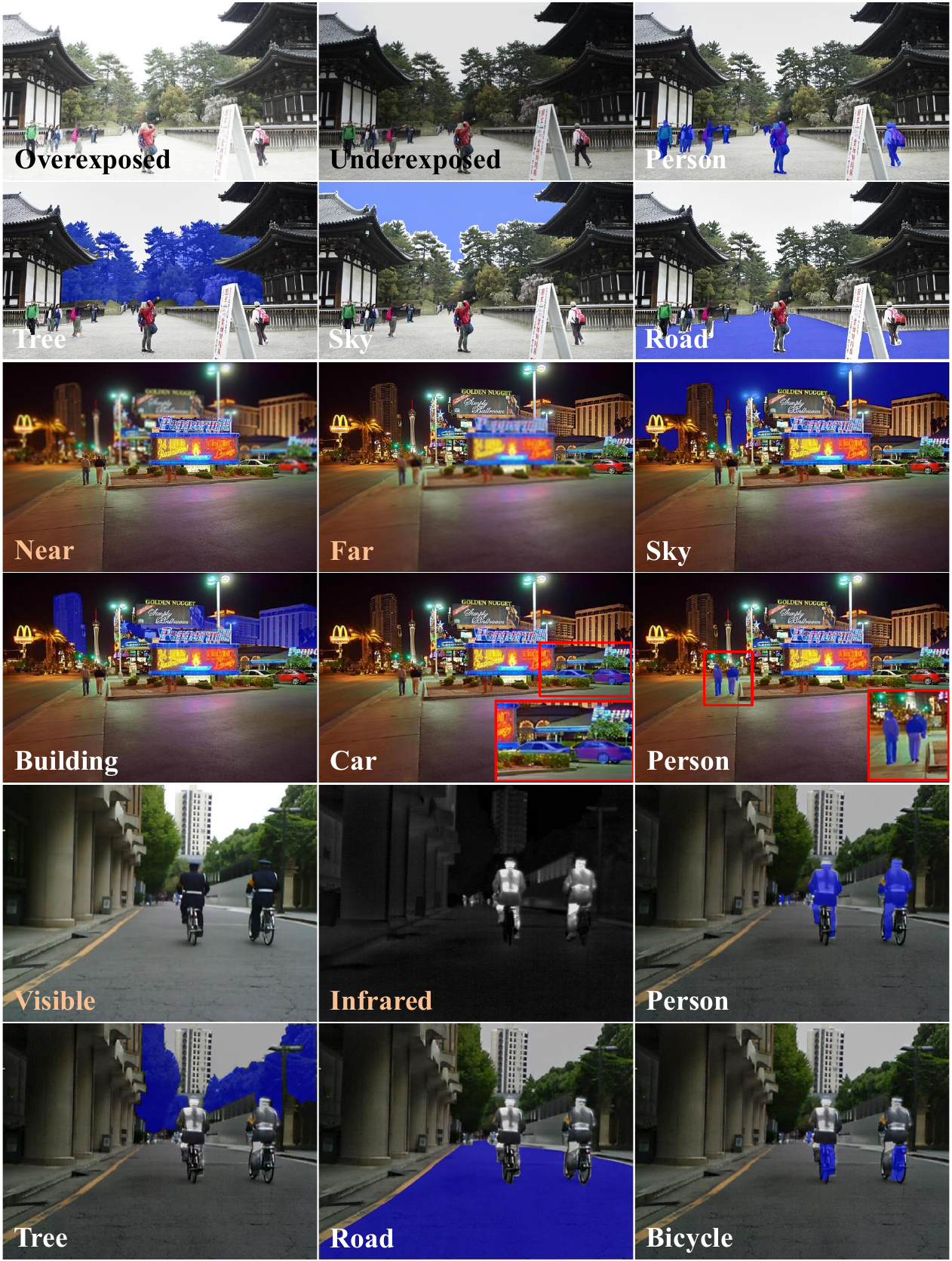} 
  \captionsetup{skip=1pt}
  \caption{Image segmentation results for different labels across various fusion-task scenarios. From top to bottom, the rows are from the SICE(MEF), MFIF(MFF), and MSRS(IVIF) datasets, respectively.}
  \vspace{-5pt}
  \label{fig:seg_result}
\end{figure}

\begin{figure}[t] 
  \centering 
  \includegraphics[width=\columnwidth]{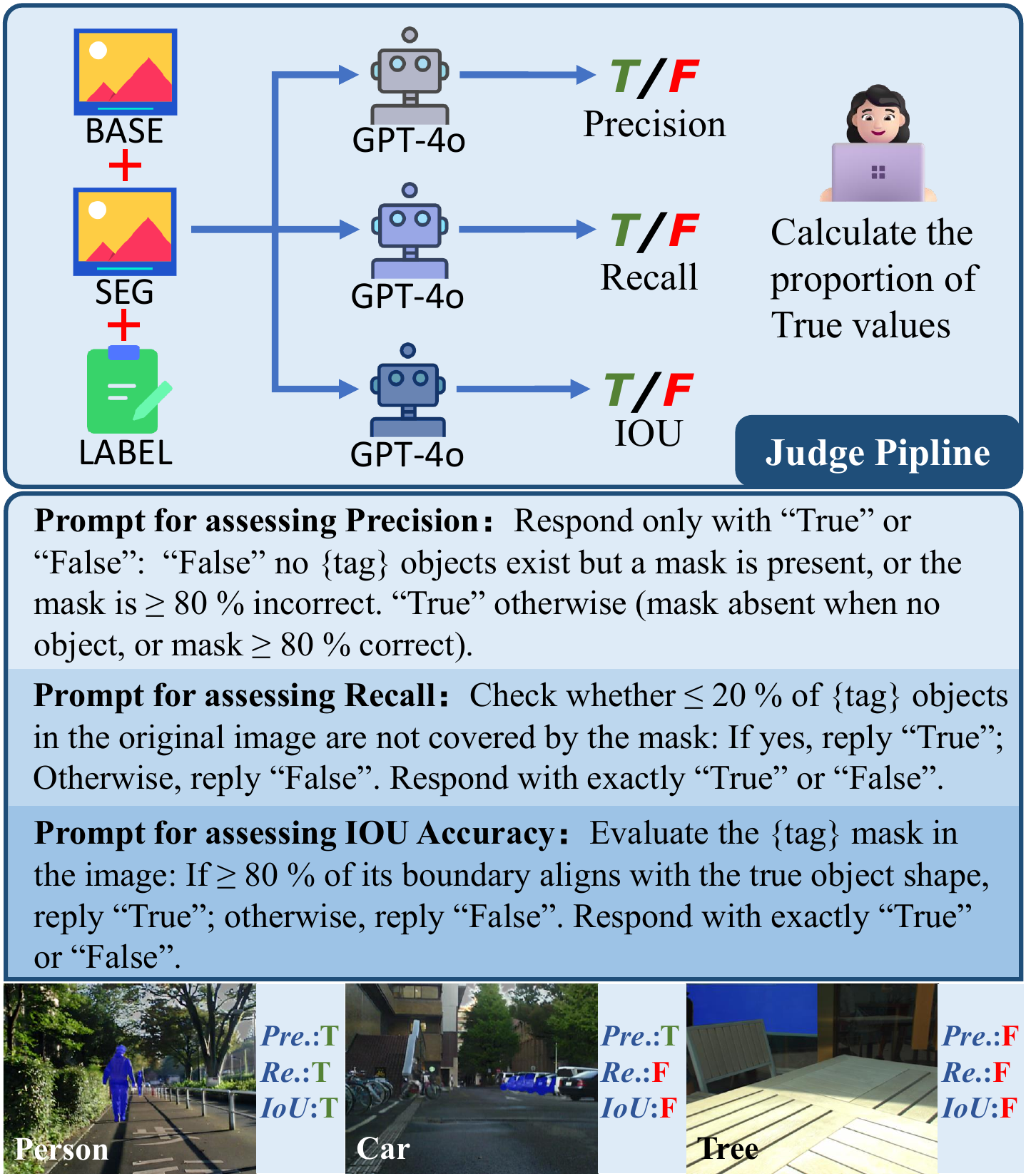} 
  \captionsetup{skip=1pt}
  \caption{The upper part is the pipeline for Segmentation Evaluation using GPT-4o. The lower part is a showcase of GPT-4o prompts and evaluation cases.}
  \vspace{-5pt}
  \label{fig:seg_judge2}
\end{figure}

A visualization of our segmentation results is shown in Fig.~\ref{fig:seg_result}. By using the segmentation training pairs constructed via our M3 synthesis method, we are able to effectively transfer segmentation capability into various fusion tasks. Furthermore, based on the selection of different labels, our method can output a segmentation map for any specified semantic region. Our work also achieves accurate results even for complex segmentation boundaries. To the best of our knowledge, this is the first work in the field of image fusion to achieve the end-to-end output of semantically controlled segmentation maps.

Here, we conduct a quantitative comparison with the open-source segmentation model (LISA~\cite{lisa}) on the FMB~\cite{liu2023multi} infrared–visible image fusion dataset, which provides segmentation ground truth, as shown in Tab.~\ref{tab:ditfuse_lisa_per_class}. Our segmentation performance in the fusion setting remains highly competitive. However, most of the fusion datasets used in our experiments do not include segmentation ground truth. How, then, can we evaluate our model’s segmentation performance in other fusion tasks and datasets?

We design a new evaluation protocol to address this. The protocol incorporates assessments from the large model GPT-4o and is supplemented with comparisons against LISA. As depicted in Fig.~\ref{fig:seg_judge2}, our process specifically involves calling the GPT-4o API and feeding it the fused image, our segmentation output, and the corresponding text label for assessment.

To ensure objective and reliable outcomes, we engineer a sophisticated, multi-dimensional evaluation system to replace the uncertainties associated with direct scoring from large models. The core of our methodology is the deconstruction of the complex evaluation task into three independent dimensions. For each dimension, rather than requesting a quantitative score, we guide GPT-4o to provide a binary ‘True/False’ judgment through carefully crafted closed-ended questions. This strategy effectively eliminates subjective bias and random fluctuations in the scoring process, establishing a structured, repeatable, and robust evaluation framework.

These three dimensions, illustrated from top to bottom in Fig.~\ref{fig:seg_judge2}, are as follows:
\begin{enumerate}
    \item \textbf{Precision:} Does the segmented region contain incorrect content? (Considered correct if errors are below 20\%).
    \item \textbf{Recall:} For a given label, is the segmented region complete? (Considered correct if over 80\% is captured).
    \item \textbf{IOU Accuracy:} For the correctly segmented area, is the contour accurate? (Considered correct if over 80\% is accurate).
\end{enumerate}

Finally, every image is assessed against these three criteria, yielding a multi-faceted evaluation result. And the examples of the corresponding evaluation results are shown in Fig.~\ref{fig:seg_judge2}.

To provide a clearer evaluation of segmentation performance across a dataset, we define three ratios. Based on our evaluation criteria, these are:
\begin{itemize}
    \item \textbf{P.Ratio:} The proportion of results within the dataset that are free of ``Precision''.
    \item \textbf{R.Ratio:} The proportion of results within the dataset that are free of ``Recall''.
    \item \textbf{I.Ratio:} The proportion of results within the dataset that exhibit correct ``IOU Accuracy''.
\end{itemize}

\begin{figure}[t] 
  \centering 
  \includegraphics[width=\columnwidth]{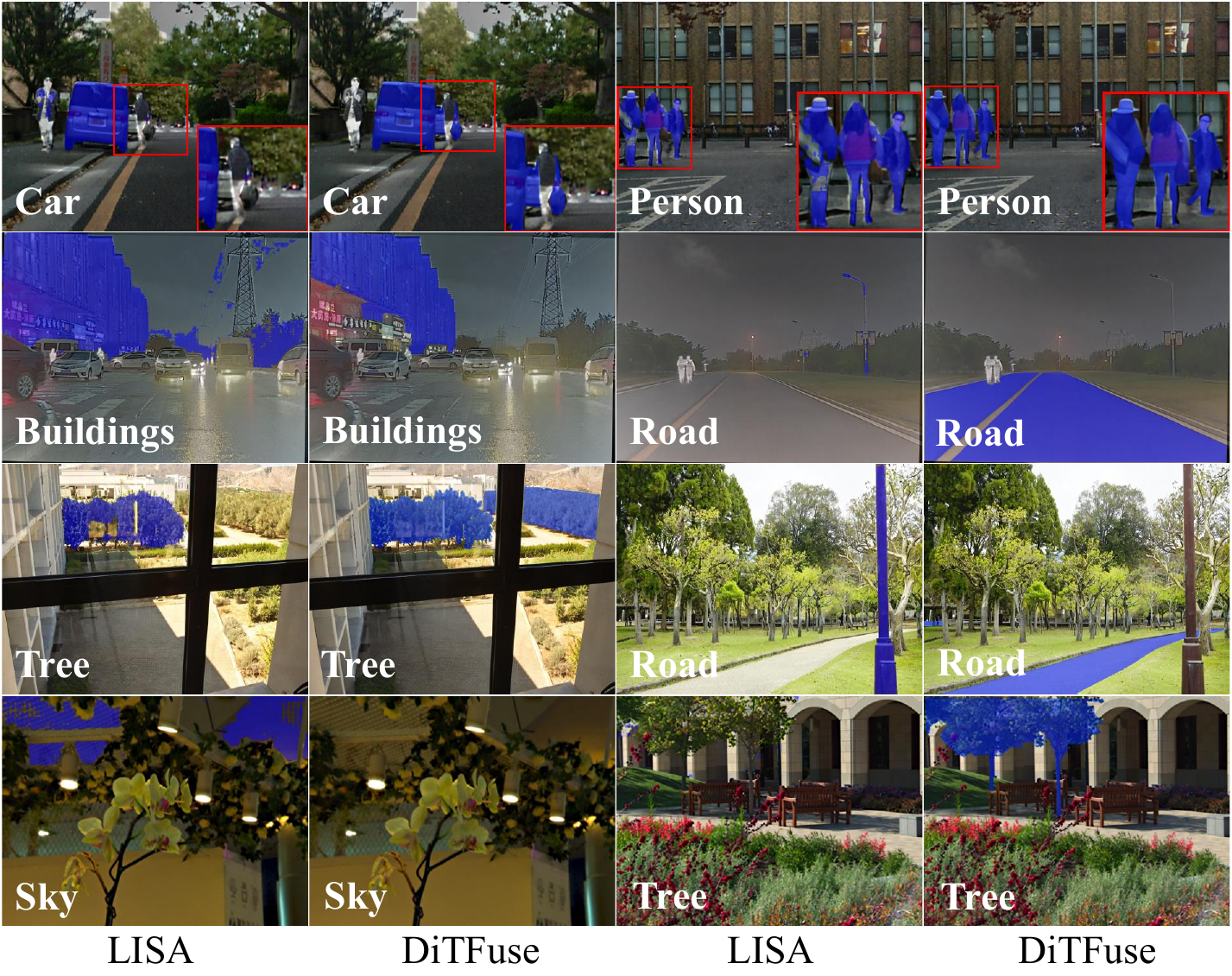} 
  \captionsetup{skip=1pt}
  \caption{Qualitative comparison of segmentation capability: LISA vs. DiTFuse}
  \vspace{-5pt}
  \label{fig:seg_cmp}
\end{figure}

Even with the aforementioned metrics, a direct comparison of our segmentation results against conventional methods is challenging. To address this, we benchmark our method against LISA~\cite{lisa}, a prominent large language model–based segmentation approach. We apply LISA to our fused images and then evaluate its outputs using the identical GPT-4o protocol, yielding three comparative performance metrics, which are presented in Tab.~\ref{tab:seg_cmp}. Our method substantially outperforms LISA. We attribute this superior performance to our training strategy involving M3-synthesized and multi-modal fusion data, which equips our model with a more sophisticated understanding of multi-modal information, leading to enhanced segmentation accuracy. Furthermore, a qualitative comparison in Fig.~\ref{fig:seg_cmp} visually confirms our method's advantages: our results exhibit more precise contours, more comprehensive coverage of target objects, and more accurate segmentation zones. Overall, these comparisons collectively demonstrate that our segmentation capability holds a significant competitive edge.

\subsection{Runtime and Efficiency Comparison for Image Fusion}

To quantify the impact of the DiT architecture on test-time speed, we compare the inference rates of existing fusion methods. For control-based methods, we measure the end-to-end pipeline time required to generate a single fused image, which includes the latency spent invoking external models for reasoning. The detailed runtime results are reported in Tab.~\ref{tab:latency_twoinone}.

\begin{table}[t]
\centering
\caption{Overview of Model Parameters and Processing Speed.}
\label{tab:latency_twoinone}
\renewcommand{\arraystretch}{0.98}
\setlength{\tabcolsep}{1.2pt} 
\footnotesize
\begin{tabular}{@{}lcc|lcc@{}} 
\Xhline{1.2pt}
\textbf{Method} & \makecell{\textbf{\#Params}\\\textbf{(M)}} & \makecell{\textbf{Latency}\\\textbf{(s/sample)}} &
\textbf{Method} & \makecell{\textbf{\#Params}\\\textbf{(M)}} & \makecell{\textbf{Latency}\\\textbf{(s/sample)}} \\
\hline
\textcolor{gray}{SwinFusion}~\cite{swinfusion}   & 0.973   & 1.290  & \textcolor{gray}{TextFusion}~\cite{TextFusion}   & 0.0739  & 0.102   \\
\textcolor{gray}{SeAFusion}~\cite{seafusion}     & 0.1669  & 1.235  & DDFM~\cite{ddfm}               & 552.66  & 34.502  \\
\textcolor{gray}{PMGI}~\cite{pmgi}               & 0.042   & 0.052  & Text-DiFuse~\cite{Text-DiFuse} & 119.455 & 80.132  \\
\textcolor{gray}{DDBFusion}~\cite{ddbfusion}     & 3.673   & 0.9256 & CCF~\cite{ccf}                 & 552.66  & 210.883 \\
\textcolor{gray}{DeFusion}~\cite{defusion}       & 7.874   & 0.086  & OmniFuse~\cite{omnifuse}       & 173.340 & 9.672   \\
\textcolor{gray}{U2Fusion}~\cite{u2fusion}       & 0.659   & 0.082  & TeRF~\cite{terf}               & 8.025\,B& 72.181  \\
\textcolor{gray}{Text-IF}~\cite{textif}          & 215.11  & 0.4492 & DiTFuse                         & 3.846\,B& 53.551  \\
\Xhline{1.2pt}
\end{tabular}
\end{table}

The methods shown in gray are non–diffusion-based, while those in black are diffusion-based. All evaluations are conducted on the same test set and resolution using a single RTX 3090 GPU. As expected, diffusion methods generally incur higher computational costs than traditional feed-forward networks. Nevertheless, our model achieves inference speed comparable to other diffusion-based approaches while maintaining a larger parameter capacity and superior fusion quality. This indicates that the proposed architecture strikes a favorable balance between generation efficiency and performance. Moreover, compared with other text-controlled methods, our end-to-end processing time per image is faster than approaches that rely on external models for control (e.g., TeRF and Text-DiFuse), further demonstrating the advantage of performing controllable generation within a single unified framework.

\subsection{Ablation Study}

We conduct comprehensive ablation studies to validate the critical components of our training strategy, including the use of task tags, data composition for fusion tasks, and the diversity of degradation types in M3.

\subsubsection{Effect of Task Tags in Training}

\begin{figure}[htbp]
  \centering
  \includegraphics[width=\linewidth]{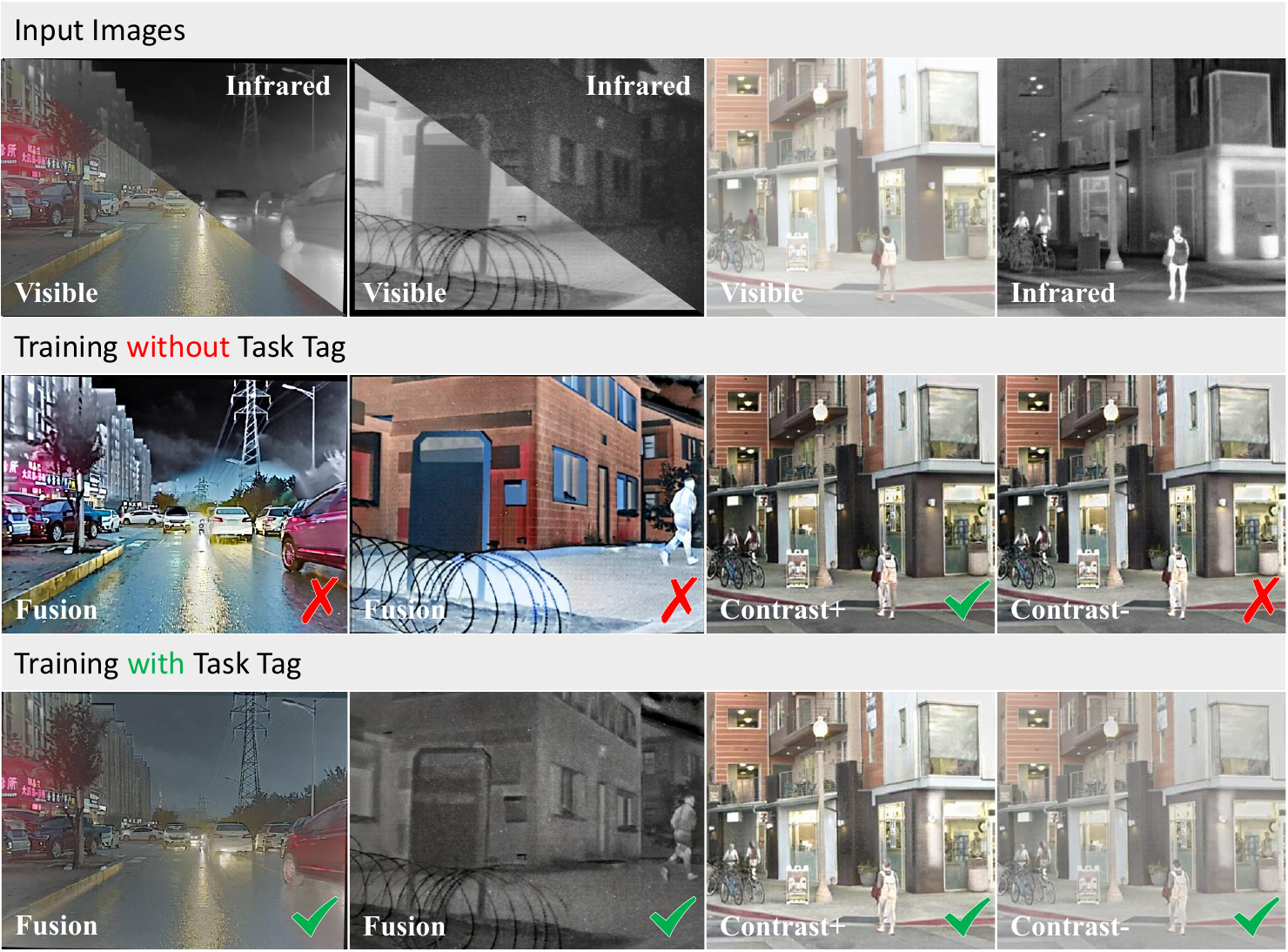}
  \captionsetup{skip=1pt}
  \caption{Ablation on the use of task tags during training. The first row shows input image pairs. The second row presents outputs from the model trained without task tags. The third row shows results from the model trained with task tags.}
  \vspace{-5pt}
  \label{fig:ab1}
\end{figure}

As shown in Fig.~\ref{fig:ab1}, we examine the effect of including explicit \texttt{[TASK]} and \texttt{<SUBTASK>} tags in the training instructions. When task tags are removed, we observe three major issues.

First, the model exhibits noticeable color distortions in standard fusion tasks (leftmost column). This occurs due to the diverse color distributions across training data, where samples from tasks such as light adjustment or segmentation differ significantly from the input images, leading to unintended color shifts.

Second, the model struggles to adhere to task constraints. In the second column, although both infrared and visible inputs from the TNO dataset are grayscale, the model generates unrealistic colorized outputs, suggesting that it applies inappropriate priors from unrelated tasks.

Third, subtask controllability deteriorates. For instance, the \texttt{<CONTRAST+>} and \texttt{<CONTRAST->} instructions both result in increased contrast (columns three and four), indicating confusion between opposing directives.

We attribute these issues to the model’s inability to distinguish between tasks with conflicting objectives. To address this, we introduce a structured tag system: task tags help separate fundamentally different goals (e.g., fusion vs. segmentation), while subtask tags guide fine-grained behavior (e.g., \texttt{<MULTI-EXPOSURE>} or \texttt{<LIGHT++>}). With these tags, the model learns stable task alignments during training and can often generalize without requiring explicit tags at inference time.

\subsubsection{Impact of Data Composition for Fusion Tasks}

\begin{figure}[t]
  \centering
  \includegraphics[width=\linewidth]{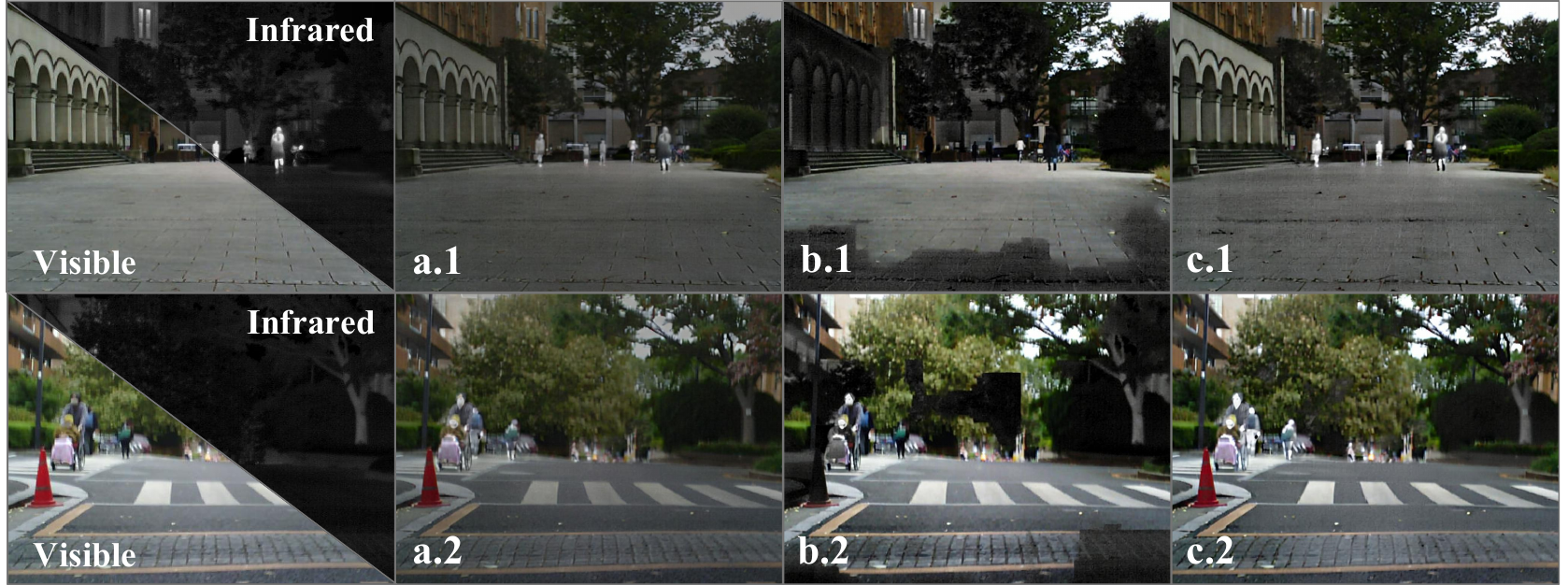}
  \captionsetup{skip=1pt}
  \caption{Ablation study on data composition within the \texttt{[FUSION]} task tag. The first column: input infrared and visible images. Column (a) results using only fusion task data. Column (b) results using only M3 data. Column (c) results using both fusion and M3 data.}
  \vspace{-5pt}
  \label{fig:ab2}
\end{figure}

We analyze how different data compositions affect fusion task performance. As illustrated in Fig.~\ref{fig:ab2}, we compare three setups under the \texttt{[FUSION]} task tag: (i) using only fusion data (e.g., IVIF, MEF, MFF), (ii) using only M3 data, and (iii) combining both.

Models trained only on fusion data tend to replicate the statistical properties of the pseudo ground truths used in those datasets, often producing overly smoothed, mean-like outputs. This suggests that directly optimizing toward such approximated targets can lead the model to converge to a suboptimal solution, rather than learning the underlying fusion objective. When trained only on M3 data, the model learns to select informative regions from multi-source inputs but lacks pixel-wise blending ability, resulting in visible block artifacts and edge inconsistencies. Combining both types of data yields the best performance. M3 teaches the model how to reason over complementary content, while fusion data enforces consistency in low-level reconstruction. This hybrid training setup produces results that are more natural, semantically rich, and visually consistent.

\begin{table}[t]
\centering
\caption{Ablation study on degradation combinations in M3. We evaluate the model on the IVIF task using the MSRS dataset under different degradation setups. The Gaussian noise–only setting corresponds to classical MIM. Our full strategy with mixed degradations achieves the best performance.}
\begin{adjustbox}{width=1.00\linewidth,center}
\begin{tabular}{ccc|cccc}
\toprule
Blur & Noise Mask & Gaussian Noise & MSE $\downarrow$ & PSNR $\uparrow$ & CLIP-IQA+ $\uparrow$ & MUSIQ $\uparrow$ \\
\midrule
\checkmark &             &               & 0.027 & 65.58 & 0.363 & 34.80 \\
           & \checkmark  &               & 0.027 & 65.48 & 0.363 & 33.64 \\
           &             & \checkmark    & 0.026 & 64.17 & 0.257 & 31.09 \\
\checkmark & \checkmark  & \checkmark    & \textbf{0.021} & \textbf{66.63} & \textbf{0.392} & \textbf{36.28} \\
\bottomrule
\end{tabular}
\end{adjustbox}
\vspace{-5pt}
\label{tab:m3_degradation}
\end{table}

\subsubsection{Effect of Degradation Diversity in M3}

To evaluate the impact of degradation diversity in M3, we experiment with four training settings: (i) blur only, (ii) noise-mask only, (iii) Gaussian-noise only, and (iv) random sampling from all three.

As shown in Tab.~\ref{tab:m3_degradation}, each individual degradation type leads to weaker performance compared to the full setting. The Gaussian noise–only configuration (MIM) particularly underperforms in perceptual metrics such as CLIP-IQA+ and MUSIQ, indicating that a single noise type is insufficient for modeling diverse degradations. In contrast, our full M3 strategy, which randomly samples among blur, noise mask, and Gaussian noise, consistently achieves the best results across all metrics. These results suggest that degradation diversity is crucial for robust feature learning and generalizable reconstruction under varied input conditions.

\subsubsection{Test-Time Prompt Performance.} Here we ablate the prompt’s influence on experimental results from three angles: (i) the effect of task tags, (ii) the performance under unseen instructions, and (iii) the impact of deliberately confusing instructions.

\begin{figure}[t] 
  \centering 
  \includegraphics[width=\columnwidth]{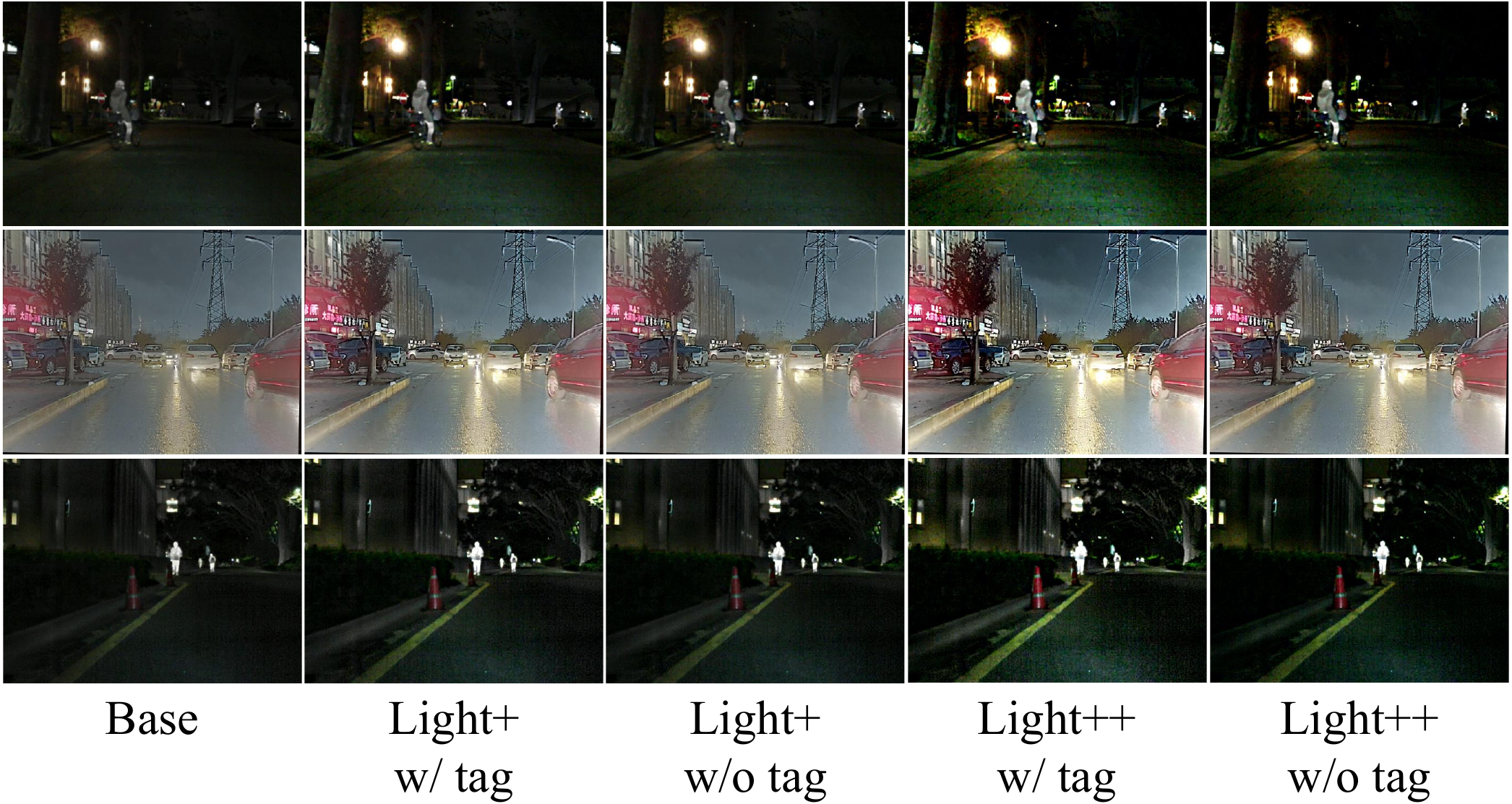} 
  \captionsetup{skip=2pt}
  \caption{Qualitative Ablation on Task Tags for Fusion.}
  \vspace{-5pt}
  \label{fig:prompt1}
\end{figure}

\textbf{Effect of Task Tags at Test Time.} As shown in Fig.~\ref{fig:prompt1}, we keep the basic template and control instruction fixed while comparing outputs with and without the task tag. We observe that removing the task tag reduces the intensity of brightness enhancement, but the overall instruction is still followed correctly. This suggests that task tags act as auxiliary semantic anchors that strengthen the intended effect, while DiTFuse remains robust enough to interpret and execute the instruction even without them.

\begin{figure}[t] 
  \centering 
  \includegraphics[width=\columnwidth]{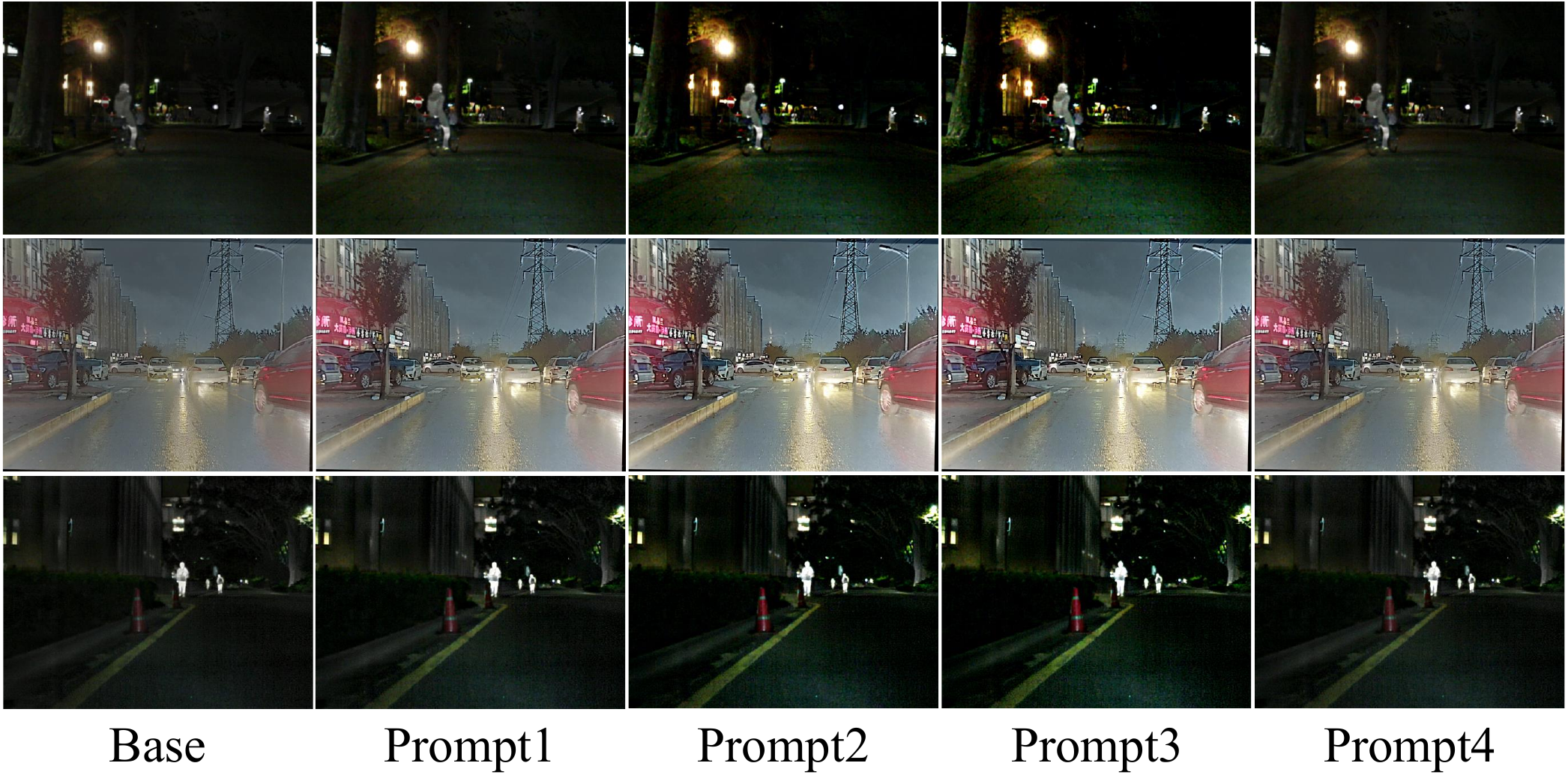} 
  \captionsetup{skip=1pt}
  \caption{Qualitative comparison with custom, unseen prompts.}
  \vspace{-5pt}
  \label{fig:prompt2}
\end{figure}

\textbf{Unseen or Simplified Prompts.} Next, we evaluate DiTFuse under custom and unseen control instructions that do not appear in the training set. To better assess the influence of the prompt itself, we remove task tags from all prompts and retain only the basic template while varying the control instruction. The prompt set is summarized in Tab.~\ref{tab:prompt2}, and qualitative results are provided in Fig.~\ref{fig:prompt2}.

\begin{table}[t]
\centering
\caption{Prompt list containing only the control instruction.}
\renewcommand{\arraystretch}{0.98}
\setlength{\tabcolsep}{1.2pt} 
\begin{tabular}{c p{0.85\linewidth}} 

\toprule 
\textbf{Prompt} & \textbf{Instruction} \\
\midrule 
Prompt1 & \texttt{\detokenize{Extract and fuse high-quality features from both images. Slightly brighten the critical elements to make them stand out subtly.}} \\
\midrule
Prompt2 & \texttt{\detokenize{Extract and fuse high-quality features from both images. Increase brightness of the input images.}} \\
\midrule
Prompt3 & \texttt{\detokenize{Fuse the images. Increase brightness of the input images.}} \\
\midrule
Prompt4 & \texttt{\detokenize{Fuse the images and increase the brightness.}} \\
\bottomrule

\end{tabular}
\vspace{-5pt}
\label{tab:prompt2}
\end{table}

In Fig.~\ref{fig:prompt2}, Base denotes the version without brightness modification. Prompt1 uses a prompt that appears in the training set, whereas Prompts2–4 are newly defined and unseen. We observe that these new prompts correctly interpret the instruction to ``increase brightness". Regardless of the instruction’s length or complexity, they all achieve a certain improvement in brightness compared with the base. This indicates that our model is capable of handling user instructions of varying complexity and maintains strong overall robustness.

\begin{figure}[t] 
  \centering 
  \includegraphics[width=\columnwidth]{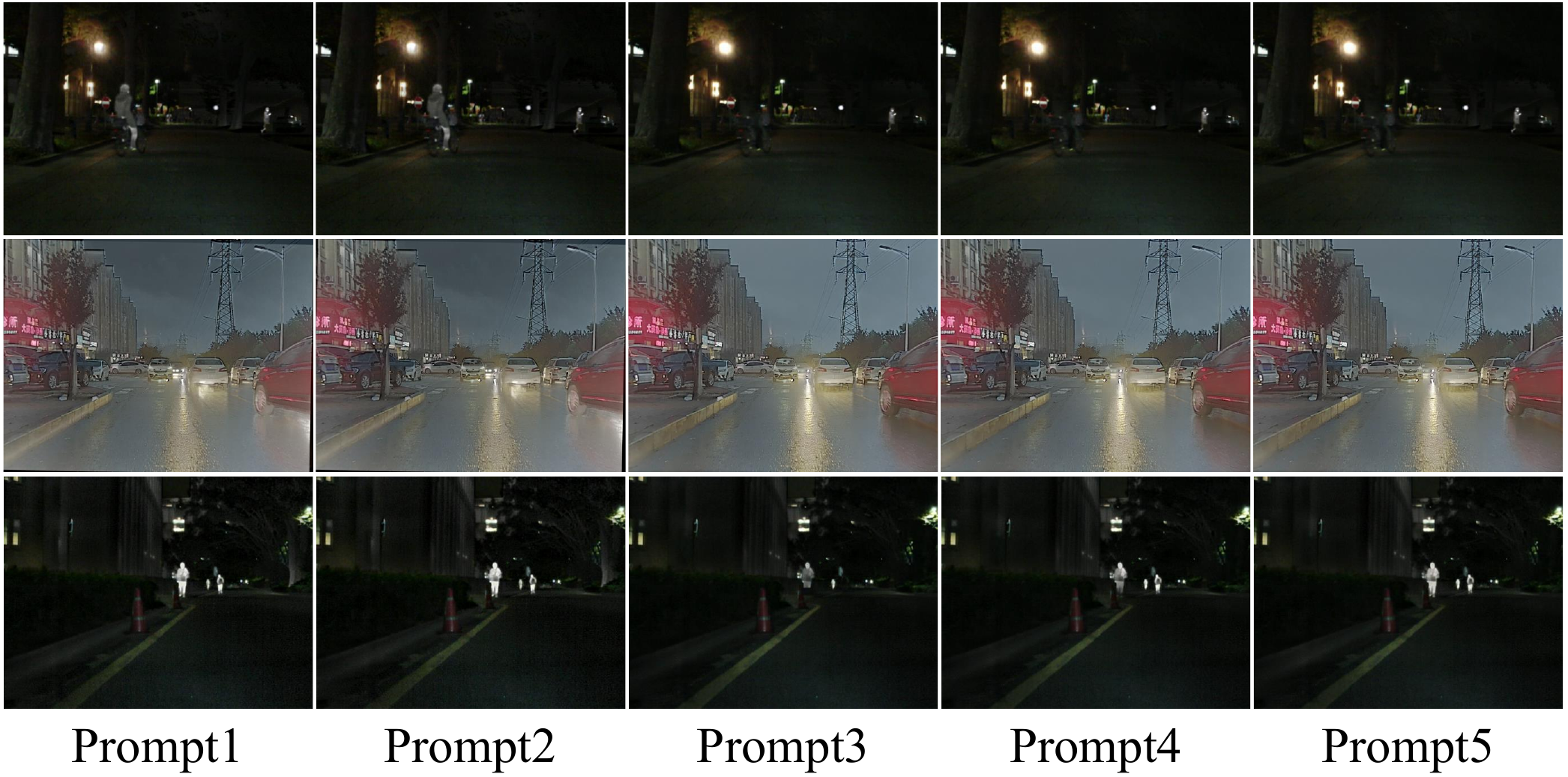} 
  \captionsetup{skip=1pt}
  \caption{Effect of modifying the basic prompt template and control instruction (task tag removed).}
  \vspace{-5pt}
  \label{fig:prompt3}
\end{figure}

\begin{table}[t]
\centering
\caption{Detailed prompt configurations corresponding to Fig.~\ref{fig:prompt3}.}
\renewcommand{\arraystretch}{0.98}
\setlength{\tabcolsep}{1.2pt} 
\begin{tabular}{c p{0.85\linewidth}} 

\toprule 
\textbf{Prompt} & \textbf{Instruction} \\
\midrule 
Prompt1 & \texttt{\detokenize{This is the first image <img><|image_1|></img>, and this is the second image <img><|image_2|></img>. Please generate the image based on the following requirements: asda324&*&*!@&^&^}} \\
\midrule
Prompt2 & \texttt{\detokenize{This is the first image <img><|image_1|></img>, and this is the second image <img><|image_2|></img>. Please generate the image based on the following requirements: fuse the images asda324&*&*!@&^&^}} \\
\midrule
Prompt3 & \texttt{\detokenize{Fuse the <img><|image_1|></img> and <img><|image_2|></img>.}} \\
\midrule
Prompt4 & \texttt{\detokenize{Fuse the <img><|image_1|></img> and <img><|image_2|></img> and seg the cars.}} \\
\midrule
Prompt5 & \texttt{\detokenize{Fuse the <img><|image_1|></img> and <img><|image_2|></img> and seg the cars and the buildings.}} \\
\bottomrule

\end{tabular}
\label{tab:prompt3}
\end{table}

\textbf{Noisy or Non-Standard Prompt Formats.} To further assess DiTFuse’s tolerance to non-standard or semantically irrelevant inputs, we modify or remove both the task tag and the basic template, and introduce prompts with unrelated or noisy text. The qualitative comparison is presented in Fig.~\ref{fig:prompt3} and Tab.~\ref{tab:prompt3}. We find that when prompts are semantically meaningless (e.g., random words or unrelated descriptions), DiTFuse still performs a reasonable baseline fusion. This robustness is largely attributed to the large-scale M3 training, which enables strong modality alignment even without explicit textual guidance. Nevertheless, removing the standardized prompt template notably degrades fusion quality and weakens controllability, since the model has been trained with that template format to ensure consistent conditioning. Therefore, performance degradation on completely unstructured inputs is expected and acceptable.

\subsection{Further Analysis}

\begin{figure}[t] 
  \centering 
  \includegraphics[width=\columnwidth]{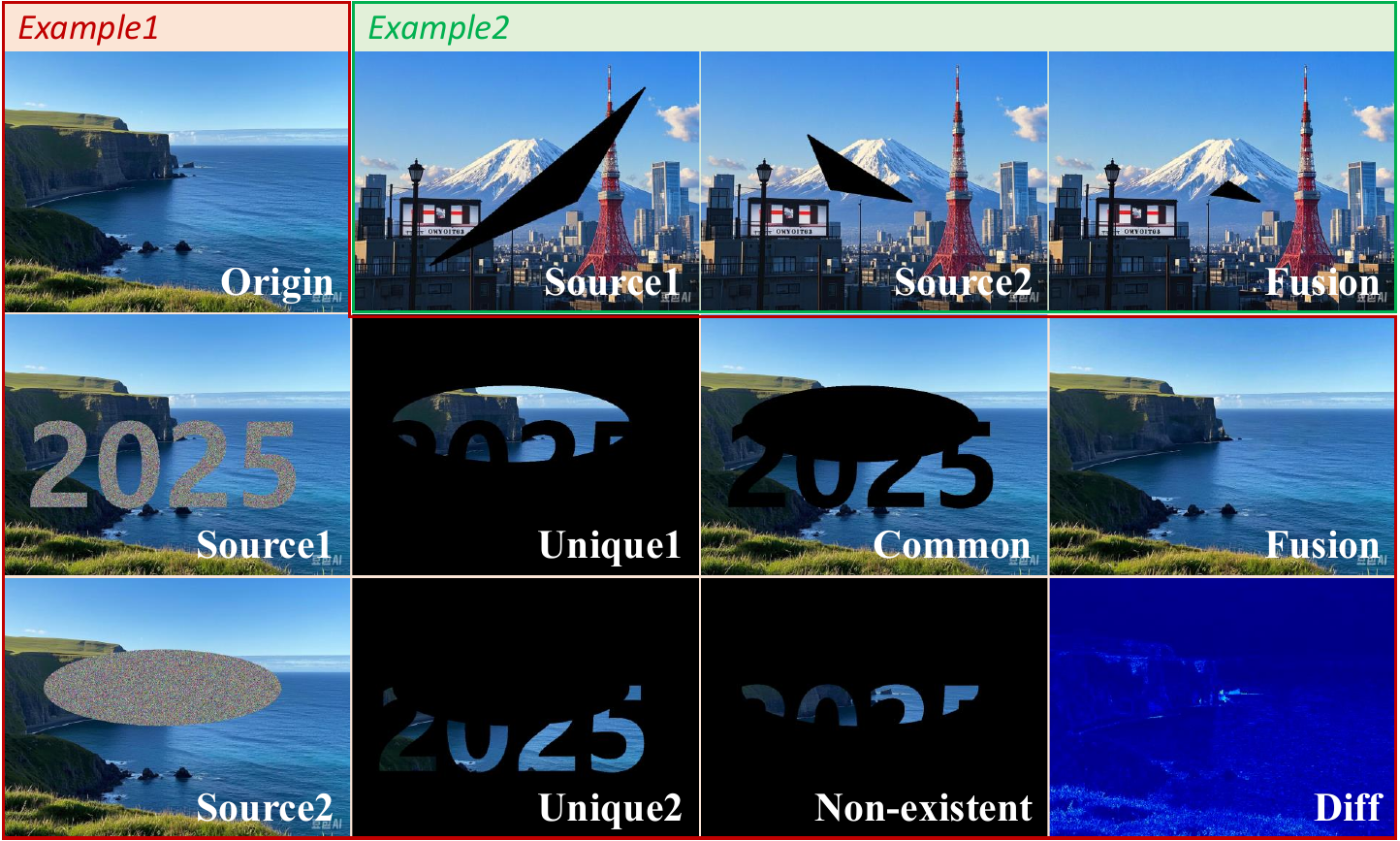} 
  \captionsetup{skip=1pt}
  \caption{In these demos, a model learns to fuse two randomly masked versions of a single image. By leveraging complementary information, it can restore the full image, even predicting content in jointly masked regions.}
  \vspace{-5pt}
  \label{fig:zero1}
\end{figure}

Furthermore, our model exhibits several zero-shot capabilities. As depicted in Fig.~\ref{fig:zero1}, two input images are generated by applying random, irregular masks to a source image. This process results in regions that are masked in one input but visible in the other, as well as regions that are jointly masked in both. The fused output demonstrates that for areas masked in only one input, the model correctly reconstructs the content by sourcing information from the corresponding unmasked area. For jointly masked regions, it generates plausible and coherent content. This capability underscores that our M3 strategy facilitates not only selective information fusion but also high-fidelity content generation, guided by a learned natural image prior.

Collectively, these experiments reveal that our model learns a robust image prior and implicit completion capability, enabling it to selectively aggregate complementary information and plausibly hallucinate missing content. This demonstrates not only strong generalization beyond seen fusion patterns, but also an inherent capacity for content-aware generation under partial observability.

\section{Conclusion}\label{sec5}

This paper introduces an instruction-controlled, all-in-one image fusion framework based on the Diffusion Transformer (DiT) architecture. This framework empowers users to generate customized fusion results that align with their specific intentions across various fusion modes. Notably, DiTFuse marks the first initiative in the fusion domain to unify image fusion with its downstream tasks, such as segmentation. It can directly perform segmentation of desired regions based on user commands, obviating the need for auxiliary networks.

The inherent structure of the DiT architecture, which explicitly separates input and output channels, effectively mitigates inter-modal redundancy and interference. This architectural advantage facilitates superior suppression of erroneous information, thereby enhancing the visual clarity and semantic richness of the fused output. Furthermore, our proposed method exhibits a degree of zero-shot capability, indicating its potential for robust generalization across diverse fusion scenarios and control tasks.

Looking ahead, as unified models continue to advance in the visual domain, DiT-based architectures have strong potential to bridge the gap between image fusion and image restoration~\cite{tang2025dspfusion, jiang2025survey}. By jointly training on large-scale, multi-task datasets, a single model could simultaneously handle denoising, deblurring, dehazing, low-light enhancement, super-resolution, and multi-modal fusion within a unified instruction space, achieving true all-in-one capability. This unified paradigm enables different tasks to share stronger natural image priors and aligned optimization directions, resulting in more robust generalization across a wide range of degradation types and multi-modal scenarios.

\noindent \textbf{Limitation.} Our current model still has room for further improvement. Compared with traditional approaches, the DiT-based image generation pipeline has a longer runtime, and there is slight detail loss during encoding. Therefore, optimizing information encoding to reduce loss and improving sampling efficiency remain key directions for future research. In addition, using GPT-4o to evaluate segmentation performance introduces some uncertainty; however, in scenarios without ground-truth labels, it serves as a reasonable compromise. Ultimately, developing a unified fusion framework that can effectively handle multiple modalities while supporting both low-level and high-level vision tasks is an important and highly promising research direction.

\ifCLASSOPTIONcaptionsoff
  \newpage
\fi

{
\bibliographystyle{IEEEtran}
\bibliography{sn-bibliography}
}

\end{document}